%% file: main.tex
\documentclass{article}

\usepackage[preprint]{corl_2026} 

\usepackage{lastpage}
\usepackage{booktabs}
\usepackage{amsmath,amssymb}
\usepackage{multirow}
\usepackage{tabularx}
\usepackage{comment}
\usepackage{subfigure}
\usepackage{caption}
\usepackage{amsmath}
\usepackage{mathtools}
\usepackage{booktabs}
\usepackage{multirow}
\usepackage{tabularx}
\usepackage{subcaption}
\usepackage{caption}
\usepackage{enumitem}
\usepackage{microtype}
\usepackage{soul}
\usepackage{verbatim}
\usepackage{xcolor}
\usepackage{pifont}
\usepackage{xspace}
\usepackage{placeins}
\hypersetup{hidelinks}

\title{Benchmarking Evidence Sufficiency in Vision--Language Models for Autonomous Driving}

\author{
  Nhat Ha~Nguyen\\
  School of Computing, College of Engineering\\
  Wichita State University 
  Wichita, KS 67260-0083, USA\\
  \texttt{b845s937@wichita.edu} \\
   \And
  Sudipta Chowdhury \\
  Department of Engineering Technology and Industrial Distribution \\
  Texas A\&M University \\
  College Station, TX 77843, USA\\
  \texttt{chowdhurys@tamu.edu} \\
   \AND
  Lokesh Das \\
  School of Computing, College of Engineering\\
  Wichita State University 
  Wichita, KS 67260-0083, USA\\
  \texttt{lokesh.das@wichita.edu} \\
}

\begin{document}
\maketitle


\begin{abstract}
    Reliable driving VLMs must do more than answer correctly or localize relevant scene actors: they must determine whether the available visual evidence is sufficient to support an answer. Existing evaluations largely treat grounding and answerability as separate measurements, making it difficult to distinguish well-supported predictions from those localized to relevant but insufficient evidence. We introduce \textbf{DriveAnswer}, a
benchmark of 20,151 questions across seven reasoning families, that operationalizes this distinction using  matched strong- and weak-evidence configurations.
Each pair preserves the underlying scene fact and semantic answer while changing
whether the observations support answering, making abstention the appropriate
response under weak evidence. On the six-family evaluation,
the strongest zero-shot baseline reaches 67.31\% strong-evidence accuracy, yet only 21.55\% of matched questions satisfy both answer-abstain behavior and complete grounding. With controlled adaptation, Gemma-4-12B reaches 63.29\% on this joint capability, nearly tripling its zero-shot performance.
Qwen3-VL-8B exhibits the same broad improvement in evidence-sensitive answering and grounding. These results show that semantic correctness,
evidence sufficiency, and grounding are distinct capabilities that should
be evaluated jointly.
\end{abstract}


\input{sections/01_introduction}
\input{sections/02_related_work}

\input{sections/03_problem}
\input{sections/method}
\input{sections/evaluation}
\input{sections/06_discussion}



\bibliography{references}  
\appendix
\input{sections/appendix_final}
\end{document}

%% file: sections/01_introduction.tex
\section{Introduction\label{sec:introduction}}
Vision--language models (VLMs) are increasingly used to interpret complex driving scenes and support perception, prediction, and planning. Hence, specialized benchmarks are increasingly necessary to systematically identify model vulnerabilities in such a critical domain. While recent benchmarks such as nuScenes-QA, DriveLM, and GraphVQA have substantially advanced the evaluation of driving VLMs, they primarily measure models' answer correctness rather than their ability to recognize and utilize required evidence when answering~\citep{qian2024nuscenesqa, sima2024drivelm,liang2021graghvqa}. This distinction is important because autonomous driving observations are constantly under partial observability: a road user may be occluded, outside the queried camera, or visible without temporal context. Under these circumstances, a model may still recover the correct label from priors or partial cues and receive QA credit. Robustness benchmarks thus show VLMs' undesirable responsiveness under weak inputs~\citep{xie2025vlms}, while selective prediction encourages abstention under uncertainty~\citep{srinivasan2024selective}.%

Motivated by this gap, we introduce DriveAnswer, a benchmark for evaluating whether VLMs know \emph{when they have seen enough to answer}. Crucially, we do not treat visual grounding as a proxy for evidence sufficiency. A model may correctly localize the relevant road user or scene region while still lacking the temporal, cross-view, or relational evidence required to support the requested inference. DriveAnswer therefore evaluates whether models can identify relevant evidence, determine whether that evidence is sufficient, and answer or abstain accordingly.%

\textbf{Contributions.}
We make four contributions: (1) an evidence-sufficient driving evaluation protocol in which correctness depends not only on the semantic answer but also on whether the admitted observations support producing it; (2) a 20,151-question training and benchmark dataset spanning seven reasoning families, with matched strong-weak evidence interventions that preserve the underlying scene fact and question while changing answer sufficiency; (3) evaluation metrics that separately expose semantic answering error, unsupported commitment, unnecessary abstention, and failures to localize the required visual evidence; and (4) a controlled staged adaptation that separately studies answer sufficiency, grounding, and explicit matched-pair supervision. The adaptation results show  substantial gains in evidence-sensitive answering, while also revealing that these capabilities respond differently to the corresponding supervision signals.%

%% file: sections/02_related_work.tex
\section{Related Work}
\textbf{Driving vision--language reasoning.}
Autonomous-driving VQA benchmarks increasingly evaluate capabilities beyond
captioning and object recognition~\citep{nie2024reason2drive, tian2025nuscenes, marcu2024eccv-lingoqa}. nuScenes-QA introduces large-scale
scene-graph-derived driving QA spanning recognition, attributes, counting, and
spatial relations~\citep{qian2024nuscenesqa}, while DriveLM organizes
perception, prediction, and planning through GraphVQA~\citep{sima2024drivelm}. More recent benchmarks broaden evaluation through
fine-grained task taxonomies, multi-view and temporal reasoning, robustness,
and cross-region generalization~\citep{li2025fine,drivespatial2026,scenepilot2026,xie2025vlms,
wang2025omnidrive,park2025nuplanqa,fruhwirth2026stsbench,cannons2025segments}.
Scene
graphs have also been used to structure visual reasoning in general VQA~\citep{hudson2019gqa, liang2021graghvqa}. In DriveAnswer, however, the graph is privileged
construction information and is never exposed to evaluated models. DriveAnswer, therefore,
is complementary to these efforts: rather than expanding the task taxonomy, it
constructs matched versions of the same question with and without the visual
evidence required to justify its answer.

\textbf{Reliability, grounding, and selective prediction.}
Correct answers do not necessarily indicate reliable visual reasoning~\citep{reich2023measuring}.
Visually grounded VideoQA shows that strong QA performance can coexist with
weak localization of the evidence supporting an answer~\citep{xiao2024can, di2024grounded}, while DriveBench finds that driving VLMs may continue
producing plausible responses from textual cues or general knowledge when
visual inputs are degraded or absent~\citep{xie2025vlms}. Reliable VQA and
selective-prediction methods address this problem by allowing models to
abstain when predictions are insufficiently reliable~\citep{whitehead2022reliable, srinivasan2024selective}. Related multimodal reliability benchmarks study complementary failures including
unanswerable visual questions, abstention under insufficient evidence, and
hallucination despite visual grounding~\citep{guo2024unk,madhusudhan2026knowing,zhu2026mohobench,
moratelli2026benchmarking,li2023evaluating,guan2024hallusionbench,
geigle2024does}. Driving-safety
benchmarks further evaluate robustness and safety-critical behavior under
broader failure conditions~\citep{li2025fine,liao2025robodrivevlm,meng2025your}.
DriveAnswer differs by defining abstention from \emph{question-specific
evidence availability}: the underlying question and scene fact are preserved
while the observations needed to support the answer are intervened upon.
Thus, a model may produce the correct semantic answer yet still fail if the
available observations do not justify answering. Table~\ref{tab:benchmark_comparison} presents a comparison between DriveAnswer benchmark and the existing ones.

\begin{table}[!ht]
\centering
\scriptsize
\caption{
Comparison with representative autonomous-driving VLM benchmarks.
``\checkmark'' denotes explicit evaluation, ``Limited'' partial coverage,
and ``--'' no primary evaluation.
}
\label{tab:benchmark_comparison}
\setlength{\tabcolsep}{4.5pt}

\begin{tabular}{@{}lccccc@{}}
\toprule
\textbf{Benchmark}
& \textbf{QA}
& \shortstack{\textbf{Multi-view /}\\\textbf{Temporal}}
& \textbf{Grounding}
& \textbf{Decision}
& \shortstack{\textbf{Matched Evidence /}\\\textbf{Sufficiency}} \\
\midrule

nuScenes-QA~\citep{qian2024nuscenesqa}
& \checkmark & Limited & -- & -- & -- \\

DriveLM~\citep{sima2024drivelm}
& \checkmark & \checkmark & Limited & \checkmark & -- \\

DriveBench~\citep{xie2025vlms}
& \checkmark & Limited & \checkmark & \checkmark & -- \\

VLADBench~\citep{li2025fine}
& \checkmark & \checkmark & Limited & \checkmark & -- \\

DriveSpatial~\citep{drivespatial2026}
& \checkmark & \checkmark & \checkmark & -- & -- \\

ScenePilot-Bench~\citep{scenepilot2026}
& \checkmark & \checkmark & Limited & \checkmark & -- \\

\textbf{DriveAnswer (ours)}
& \checkmark
& \checkmark
& \checkmark
& \checkmark
& \textbf{\checkmark} \\

\bottomrule
\end{tabular}
\end{table}

%% file: sections/03_problem.tex
\section{Problem Formulation}
\label{sec:problem}
Let $S$ denote a driving scene, $q$ a grounded question, and $c$ an observation configuration. The benchmark constructor has access to privileged world state $W(S)$ from source annotations, calibrated geometry, persistent tracks, and semantic maps, but  evaluated models do not. Their model-facing input is
\begin{equation}
    X_{q,c}=\bigl(q,\mathcal{O}_c(S)\bigr),
\end{equation}
where $\mathcal{O}_c(S)$ contains only the unboxed camera views and time steps and camera identities as well as temporal indices associated with each observation as detailed by configuration $c$.%

Each question has a semantic reference answer $z_q^{\star}$ and a set of required evidence units $\mathcal{R}_q=\{e_1,\ldots,e_m\}$ (see Appendix~\ref{app:evidence_definitions} for formal definitions). These units represent the visual support needed to justify the answer, such as grounding a queried road user, establishing a spatial or map relation, maintaining identity across time, or observing the temporal context required to infer motion. Under a particular configuration $c$, the evidence unit may either be labeled as supported, partially available, ambiguous, or unavailable to distinguish visually difficult cases and prioritize manual verification (see Appendix~\ref{app:sufficiency_definition} for formal definitions). Clear supported or unavailable cases can be assigned directly, while partial or ambiguous cases are routed for manual verification. These evidence-unit states are then aggregated into a configuration-level sufficiency label. A configuration is sufficient only when the admitted observations collectively support every mandatory evidence unit required by the question. If one or more mandatory units is not labeled supported, the configuration is considered insufficient. 

For each question, DriveAnswer constructs a matched strong-weak pair $(c^+,c^-)$, such that 
$X^+ = X_{q,c^+}, \hspace{2pt} X^- = X_{q,c^-}$. The \emph{strong} configuration $X^+$ preserves the required visual evidence
to justify the semantic answer, whereas the matched weak configuration $X^-$ removes
critical evidence while preserving the
same underlying scene fact and question $q$. Consequently, even though the
underlying fact is unchanged,
the desired model output is different: under $X^+$ the model should
produce $z_q^\star$, while under $X^-$ it should select \emph{abstention} as the  answer.
Thus, the benchmark separates knowing the semantic answer from whether the currently
available observations justify producing it. For a matched pair $(c^+,c^-)$, success requires both $\hat z_{q,c^+}=z_q^{\star} \text{ and } \hat z_{q,c^-}=\text{ABSTAIN}$. Additional construction checks verifying that matched strong-weak pairs have the same underlying question and world-state target are provided in Appendix~\ref{app:pair_integrity}.

%% file: sections/method.tex
\section{Method}
\label{sec:new_method}
\subsection{Benchmark Construction Overview}
\label{sec:benchmark_construction}
We separately construct \emph{scene truth}, \emph{observation-level visual support}, and \emph{question-specific evidence requirements} to create matched strong and weak configurations while preserving the underlying scene fact and question semantics. The construction pipeline proceeds from a query-independent scene representation to question-specific evidence configurations. For each eligible
scene fact, we: (i) identify a verified world-state fact, (ii) generate a question targeting that fact, (iii) recover the semantic reference answer independently, (iv) specify the mandatory evidence units required to support the answer, (v) identify the camera observations and grounding targets that instantiate those units, (vi) determine whether each required evidence unit is supported by the admitted
observations, (vii) aggregate those support states into a configuration-level sufficiency
label, (viii) construct matched strong and weak configurations, and (ix) apply automatic integrity checks followed by targeted human review where needed.
\subsubsection{Layered Representation}
DriveAnswer uses four linked representations to keep semantic truth, visual
observability, and question-specific evidence requirements distinct.%

\textbf{Layer 1: query-independent scene representation.}
The first layer describes the observable driving scene independently of any
particular benchmark question. It provides the scene-level entities, temporal
context, camera observations, and identifiers needed to construct candidate
examples without conditioning the representation on a downstream query.%

\textbf{Layer 2: privileged world-state graph.}
The second layer encodes verified entities, attributes, and relations in the
underlying scene. This layer is used to establish semantic scene truth and to
identify eligible facts from which questions can be constructed. Graphs are hidden privileged entities.

\textbf{Layer 3: observation-level evidence candidates.}
The third layer links relevant world-state entities and relations to the camera observations in which they may be visually supported. It records projection and observation provenance together with cues such as visibility, truncation, occlusion, temporal continuity, and synchronized cross-view support. These signals identify candidate evidence and difficult observation conditions; they do not determine whether a question is answerable. However, a valid projection does not establish that the queried fact is recoverable from the observation. Thus, world-state verification determines \emph{what is true}, whereas question-specific evidence determines whether that truth can be justified from the observations supplied to the model.%

\textbf{Layer 4: question-specific evidence and matched configurations.}
The final layer specifies the mandatory evidence units required to answer a
particular question. Each required unit is associated with its relevant
observations and grounding targets, and each admitted configuration is assigned
a support state. A configuration is sufficient only when all mandatory evidence
units are supported. Strong configurations retain sufficient evidence, whereas
matched weak configurations remove or withhold at least one mandatory unit
while preserving the same underlying scene fact and question.

\begin{figure}[!ht]
    \centering
    \includegraphics[width=\textwidth]{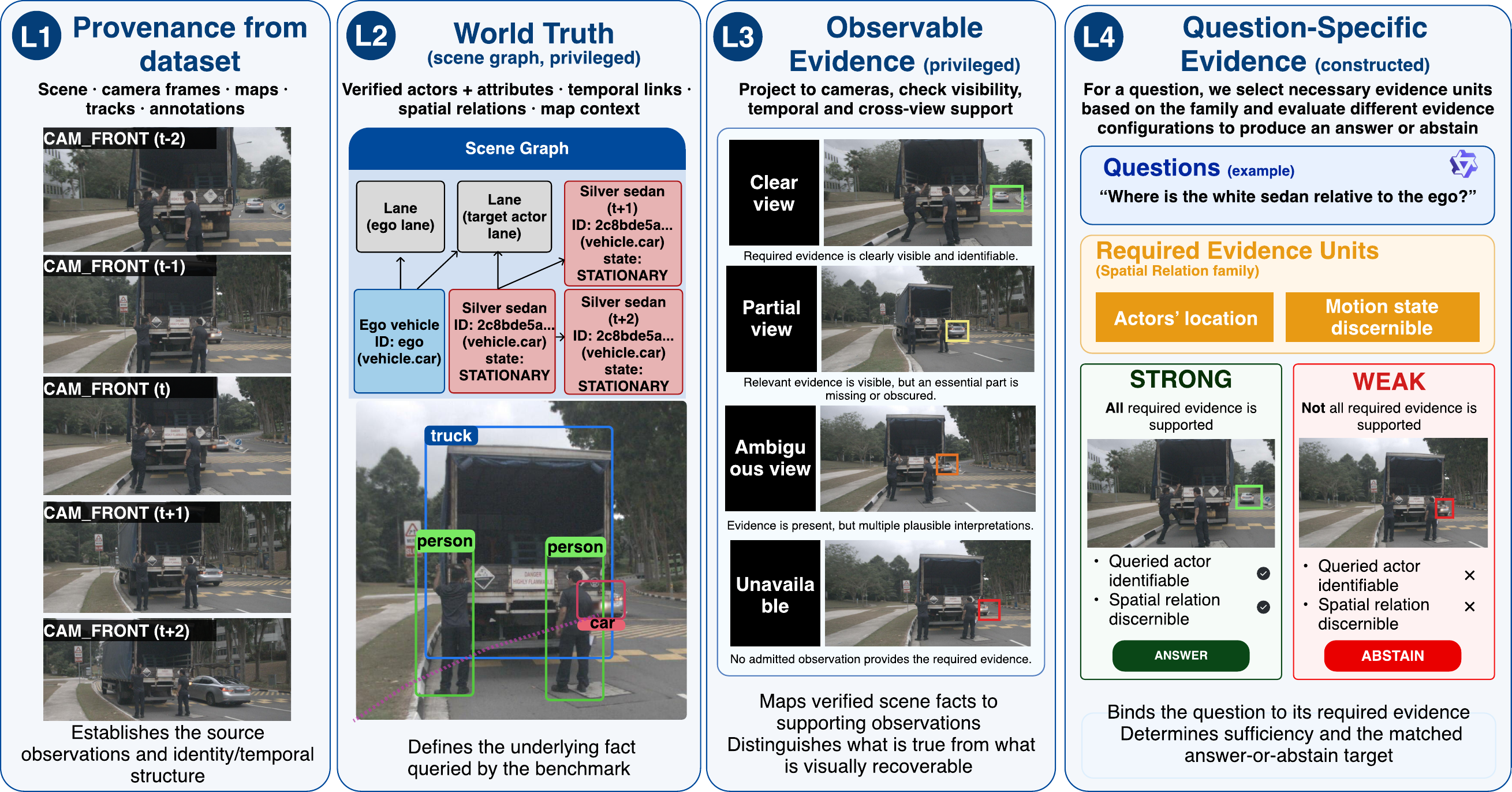}
    \caption{Overview of DriveAnswer. Layers 1--3 encode provenance, privileged world truth, and observation-level evidence. Layer~4 defines questions, required evidence, sufficiency, and answer-or-abstain target. Matched configurations vary the evidence availability while preserving
    the underlying scene fact and question. Models receive only the question and original unboxed observations; privileged construction records are withheld. Boxes are illustrative only. All images are original nuScenes~\citep{caesar2020nuscenes} frames, without blur, degradation, or other image quality modification.
    }
    \label{fig:overview}
\end{figure}

Figure~\ref{fig:overview} summarizes these four layers and their role in the
construction pipeline. Additional schema and implementation details are
provided in the Appendix \ref{app:expert_protocol}.
\subsubsection{Question and Evidence Construction}
\label{sec:question_and_evidence_construction}
For each eligible world-state fact, the benchmark constructs a question targeting that fact, while the semantic reference answer is recovered independently from the verified scene representation. We employ Qwen3-VL-8B~\citep{bai2025qwen3vl} to assist question generation using a predefined template. Qwen is used only to verbalize questions and referring expressions; reference answers and sufficiency labels are derived independently from verified construction records. Qwen proposes natural-language referring expressions from the construction-only target regions (e.g., object category, appearance, and coarse
image position). These generated attributes are not treated as semantic ground
truth, but as candidates. A candidate is retained only if the intended target is visually identifiable,
the description is unique in the admitted observation, same-category
distractors are accounted for, and the wording does not reveal the queried
relation or reference answer. Descriptions containing unsupported or unstable
attributes are rewritten or rejected. We then specify the mandatory evidence units required to justify that answer. These requirements are family specific and are defined in Appendix~\ref{app:evidence_definitions}. These constructions are separate from model evaluation. The benchmarked model receives the
original unboxed observation; construction-only regions are never exposed at
evaluation time. 

DriveAnswer spans seven reasoning families. The six canonical scene-understanding
families are \emph{Spatial Relation}, \emph{Map Relation}, \emph{Motion State},
\emph{Temporal Relation}, \emph{Temporal Recovery}, and
\emph{Cross-View Recovery}. Together they test whether the model can answer questions with spatial, map, motion,
temporal, and multi-view scene properties. \emph{Evidence-Conditioned Decision} extends the same
matched-evidence framework to a decision policy and is evaluated separately as a policy-conditioned extension rather than as
online driving-policy evaluation. The Evidence-conditioned Decision construction details are provided in Appendix~\ref{app:decision}.
\subsubsection{Review and Reconciliation}
\textbf{Automatic Validation.}
Before an example is admitted to the frozen benchmark, it passes a sequence of
automatic consistency checks covering the main construction joins. These checks verify that:
(i) the question and reference answer correspond to the same verified
world-state fact;
(ii) all required evidence units are represented in the question-specific
annotation;
(iii) grounding targets resolve to valid admitted observations;
(iv) strong configurations contain support for every mandatory evidence unit;
(v) weak configurations remove support for at least one mandatory unit;
(vi) the strong and weak examples preserve the same underlying question and
semantic fact; and
(vii) no configuration is admitted with unresolved structural or grounding
inconsistencies. The audit is deliberately conservative. Failed joins, invalid
grounding, or internally inconsistent support records are repaired when the
underlying evidence can be verified; otherwise, the affected pair is
rejected.

\textbf{Human Validation.}
Automatic checks cannot fully resolve cases in which visual support is
borderline, multiple interpretations remain plausible, or annotation sources
disagree. Such cases are routed to targeted manual review. Review focuses on three questions:
(1) whether the reference answer is supported by the verified scene semantics,
(2) whether the designated observations actually support each mandatory
evidence unit, and
(3) whether the resulting strong-weak sufficiency distinction is unambiguous. When annotation components disagree, we reconcile them manually so that
a semantic disagreement is resolved at the world-state level; and an
observation-support disagreement is resolved by reviewing the admitted visual
evidence and the family-specific evidence requirement. Cases that remain genuinely ambiguous after reconciliation are excluded to prevent annotation uncertainty, while difficult but solvable cases are retained. We additionally perform benchmark-level audits over graph construction,
projection coverage, evidence reconstruction, support consistency, and sampled
human validation (see Appendices \ref{app:human_validation_protocol} and \ref{app:annotation_qc}).

\subsection{Controlled Adaptation Study}
\label{sec:adaption}
The primary contribution of this work is the DriveAnswer benchmark. We use a
controlled adaptation study with Gemma-4-12B and Qwen3-VL-8B to test how different forms of supervision address failures exposed by the benchmark. Full implementation details are provided in Appendix~\ref{app:adapted}.

\textbf{Training setup.}
All reported adaptation stages start from the same fresh Gemma base rather than from the previous stage. This staged adaptation allows us to examine the effects of each component on answering and grounding capabilities of the resulting models. For our adaptation, we use LoRA with rank~8 and $\alpha=16$ adapting the query and value projections of the attention layers, BF16, gradient checkpointing, physical batch size~1, a maximum image side of 384 pixels with ratio preserved, and AdamW. The learning rate for the supervised fine-tuning runs is $3\times10^{-5}$. A feasibility run fits at approximately 29~GiB of VRAM on an RTX~5090. The training and validation splits contain 14,084 and 3,076 matched questions for seven-family and 13,215 train and 2,974 validation matched questions for the main six-family test, respectively. Stages B-D objectives are applied unchanged to the Qwen adaptation.

\textbf{Stage A: Zero-shot baseline.}
Stage A evaluates the unadapted pretrained backbone and serves as the reference point for subsequent stages. It appears as zero-shot in tables and evaluation.

\textbf{Stage B: Answer + Sufficiency SFT.} 
Each matched question contributes its strong configuration with the target $z_q^\star$ and its weak configuration with target \text{ABSTAIN}. The per-pair answer objective is the mean of the two independent cross-entropy terms,
\begin{equation}
    \mathcal{L}_{\mathrm{ans+suff}}=\tfrac{1}{2}\left[
    \operatorname{CE}\bigl(f(X^+),z_q^\star\bigr)+
    \operatorname{CE}\bigl(f(X^-),\text{ABSTAIN}\bigr)\right].
\end{equation}
The two conditions are supervised jointly as a matched training unit but optimized through independent cross-entropy terms. Stage B introduces no visual-grounding loss nor explicit cross-pair coupling, isolating evidence-sufficiency supervision.

\textbf{Stage C: + Grounding supervision.}
Stage C retains Stage B and adds supervised grounding. Grounding tasks are averaged before being added to the objective, preventing questions with more grounding targets from receiving greater training weight. 
The implemented Stage C objective is%
\begin{equation}
    \mathcal{L}_{\mathrm{ground+evid}}
    =
    \mathcal{L}_{\mathrm{ans+suff}}
    +
    \lambda_g \mathcal{L}_{\mathrm{ground}},
\end{equation}%
where $\lambda_g$ is 0.5 and $\mathcal{L}_{\mathrm{ground}}$ is the mean supervised loss over the grounding tasks associated with that matched pair. 

\textbf{Stage D: + Explicit matched-pair objective.}
Stage D retains Stage C  and
adds a ranking loss that couples the two matched evidence conditions.
The pair term encourages the semantic answer under $X^{+}$ and
\textsc{ABSTAIN} under $X^{-}$ while also encouraging the corresponding change
in relative model support across the matched intervention. The complete
objective is%
\begin{equation}
    \mathcal{L}_{\mathrm{pair\text{-}aware}}
    =
    \mathcal{L}_{\mathrm{ans+suff}}
    +
    \lambda_g \mathcal{L}_{\mathrm{ground}}
    +
    \lambda_m \mathcal{L}_{\mathrm{match}},
\end{equation}%
with $\lambda_m=0.25$ and $\lambda_g=0.5$. This stage tests whether direct coupling across strong-weak predictions provides
additional benefit. We detail the full training objective of Stage~D in Appendix~\ref{app:stageD_training}.%

%% file: sections/evaluation.tex
\section{Experiments}
\label{sec:evaluation}

\subsection{Evaluation Protocol}

The dataset is split by scene into 739 training, 25 validation, and
38 test scenes with no overlap. The released
seven-family benchmark contains
2,991 matched questions; our main
evaluation and controlled adaptation study use the six
scene-understanding families with 2,830 matched questions.
Evidence-Conditioned Decision and the 7-family results are reported in Appendices~\ref{app:decision} and~\ref{app:7fam_adaptation}.

DriveAnswer reports complementary metrics designed around the matched
strong--weak evidence intervention. \textbf{Strong Semantic Accuracy (Strong)}
measures answer correctness when sufficient evidence is available, whereas
\textbf{Weak Abstention Accuracy (Weak)} measures whether the model abstains
after critical evidence is removed. Our primary evidence-sufficiency
metric, \textbf{Matched-Pair Success (Pair)}, requires both behaviors for the
same question. \textbf{Unsupported Commitment Rate (UCR)}
measures answering under insufficient evidence, while
\textbf{Unnecessary Abstention Rate (UAR)} measures abstaining despite
sufficient evidence.

Answer behavior alone does not establish whether a model can identify the
visual evidence supporting its prediction, highlighting the importance of grounding evaluation. A predicted region is considered correctly localized when its intersection-over-union with the reference region satisfies IoU $\geq t$. Using this threshold, we measure evidence-presence accuracy (\textbf{Presence}), the ratio of correcty grounded target at $t$ (\textbf{IoU@$t$}), and joint presence-localization success (\textbf{Joint@$t$}). Because a question may require multiple grounding targets, \textbf{All-G@$t$} requires every target associated with the question to achieve Joint@$t$ success.
\textbf{Joint Answer--Grounding Success (JAGS@$t$)} additionally requires Pair Success together with correct grounding of all required evidence.
These metrics expose complementary failure modes in semantic answering,
evidence sufficiency, and grounding.

\subsection{Construction Validity and Sufficiency Controls}

Automated checks found no invalid references, duplicate node identifiers, broken temporal links, or inconsistencies between questions and observation configurations. Each retained strong-weak pair preserves the underlying scene fact and semantic
question while changing only the available evidence. Structural consistency alone does not establish visual sufficiency, since many queried objects remain visible while the information required to justify the answer, such as motion history, temporal
change, cross-view context, or relational support, is removed. 
Full construction statistics, audit, visibility, and shortcut audits are reported in Appendices~\ref{app:graph_validation},~\ref{app:statistics} and~\ref{app:shortcut_controls}.

\subsection{Zero-Shot Model Performance}

We evaluate Gemma-4-12B~\citep{gemma4_2026}, Qwen3-VL-8B~\citep{bai2025qwen3vl}, InternVL3.5-8B~\citep{wang2025internvl3}, Molmo2-ER~\citep{clark2026molmo2}, and
LLaVA-OneVision-7B~\citep{llavaonevision_2024} under the same six-family held-out protocol.

\begin{table}[!ht]
\scriptsize
\centering
\caption{
Zero-shot performance on the six-family test set at $t=0.50$.
}
\label{tab:zeroshot_main}
\small
\setlength{\tabcolsep}{4.5pt}

\textbf{(a) Answer / Abstention}

\vspace{10pt}
\begin{tabular}{@{}lccccc@{}}
\toprule
\textbf{Model}
& \textbf{Strong (\%) $\uparrow$}
& \textbf{Weak (\%) $\uparrow$}
& \textbf{Pair (\%) $\uparrow$}
& \textbf{UCR (\%) $\downarrow$}
& \textbf{UAR (\%) $\downarrow$} \\
\midrule
Gemma-4-12B
& \textbf{67.31} & \textbf{69.47} & \textbf{44.35}
& \textbf{30.42} & \textbf{6.54} \\

Qwen3-VL-8B
& 54.81 & 49.26 & 17.42 & 50.74 & 11.77 \\

InternVL3.5-8B
& 55.44 & 44.06 & 13.71 & 55.94 & 13.18 \\

Molmo2-ER
& 53.64 & 41.02 & 14.13 & 58.98 & 17.39 \\

LLaVA-OneVision-7B
& 39.47 & 29.33 & 8.62 & 70.64 & 6.64 \\
\bottomrule
\end{tabular}

\vspace{10pt}
\scriptsize
\textbf{(b) Grounding / Joint Success}

\vspace{10pt}
\begin{tabular}{@{}lccccc@{}}
\toprule
\textbf{Model}
& \textbf{Presence (\%) $\uparrow$}
& \textbf{IoU@$t$ (\%) $\uparrow$}
& \textbf{Joint@$t$ (\%) $\uparrow$}
& \textbf{All-G@$t$ (\%) $\uparrow$}
& \textbf{JAGS@$t$ (\%) $\uparrow$} \\
\midrule
Gemma-4-12B
& \textbf{88.98} & 61.92 & 60.43 & 43.18 & \textbf{21.55} \\

Qwen3-VL-8B
& 84.22 & \textbf{68.10} & \textbf{62.08}
& \textbf{47.42} & 7.49 \\

InternVL3.5-8B
& 75.77 & 0.07 & 15.22 & 0.07 & 0.04 \\

Molmo2-ER
& 83.97 & 0.35 & 14.57 & 0.04 & 0.00 \\

LLaVA-OneVision-7B
& 75.59 & 0.05 & 15.09 & 0.00 & 0.00 \\
\bottomrule
\end{tabular}
\end{table}
\textbf{Zero-shot models fail to condition answers reliably on evidence
availability.}
Table~\ref{tab:zeroshot_main} shows a substantial gap between semantic answering
and evidence-sensitive behavior across backbones. Gemma-4-12B
achieves the strongest overall zero-shot performance, reaching 67.31\% Strong
Accuracy but only 44.35\% Pair Success. The gap is larger for the remaining
models: Qwen3-VL-8B reaches 54.81\% Strong Accuracy but only 17.42\% Pair
Success. Similarly, InternVL3.5-8B, Molmo2-ER, and LLaVA-OneVision-7B also obtain low Pair
Success. 
These failures arise from different behaviors. High UCR values, ranging from
30.42\% to 70.64\%, reveal frequent answering after  required evidence removal. Conversely, UAR values of 13.18\% (InternVL3.5-8B) and 17.39\% (Molmo2-ER) demonstrate abstaining when
sufficient evidence is available.
\begin{table}[!ht]
\centering
\small
\caption{
Per-family zero-shot performance on the canonical six-family held-out
test set at $t=0.50$. We report semantic answering (Strong), matched
evidence-sensitive answering (Pair), target-level grounding (Joint),
complete evidence recovery (All-G), and their joint success (JAGS).
Full per-family metrics are reported in Appendix~\ref{app:zero_shot}. \textbf{All results are expressed as percentages.}
}
\label{tab:zeroshot_perfamily}
\setlength{\tabcolsep}{4pt}

\begin{tabular}{@{}llccccc@{}}
\toprule
\textbf{Family} &
\textbf{Model} &
\textbf{Strong $\uparrow$} &
\textbf{Pair $\uparrow$} &
\textbf{Joint@.50 $\uparrow$} &
\textbf{All-G@.50 $\uparrow$} &
\textbf{JAGS@.50 $\uparrow$} \\
\midrule

Cross-view
& Gemma-4-12B        & 54.73 & 53.73 & 14.18 & 2.49 & 1.99 \\
& Qwen3-VL-8B        & 51.74 & 51.74 & 23.38 & 2.99 & 1.99 \\
& InternVL3.5-8B     & 60.70 & 19.90 & 0.00 & 0.00 & 0.00 \\
& Molmo2-ER           & 20.40 & 18.41 & 0.25 & 0.00 & 0.00 \\
& LLaVA-OneVision-7B & 73.13 & 32.34 & 0.00 & 0.00 & 0.00 \\
\midrule

Map
& Gemma-4-12B        & 54.59 & 51.89 & 75.41 & 54.05 & 29.73 \\
& Qwen3-VL-8B        & 43.24 & 25.95 & 78.11 & 57.30 & 15.14 \\
& InternVL3.5-8B     & 39.46 & 7.03 & 48.65 & 0.00 & 0.00 \\
& Molmo2-ER           & 41.08 & 4.32 & 49.19 & 0.54 & 0.00 \\
& LLaVA-OneVision-7B & 36.76 & 28.65 & 49.19 & 0.00 & 0.00 \\
\midrule

Motion
& Gemma-4-12B        & 76.28 & 33.90 & 61.09 & 45.72 & 18.85 \\
& Qwen3-VL-8B        & 83.23 & 1.08 & 62.98 & 52.21 & 0.18 \\
& InternVL3.5-8B     & 82.96 & 2.98 & 0.00 & 0.00 & 0.00 \\
& Molmo2-ER           & 80.25 & 0.54 & 0.18 & 0.00 & 0.00 \\
& LLaVA-OneVision-7B & 46.62 & 0.09 & 0.00 & 0.00 & 0.00 \\
\midrule

Spatial
& Gemma-4-12B        & 69.70 & 51.27 & 80.09 & 64.63 & 36.87 \\
& Qwen3-VL-8B        & 31.22 & 27.19 & 83.13 & 70.39 & 19.82 \\
& InternVL3.5-8B     & 34.56 & 22.24 & 45.63 & 0.23 & 0.12 \\
& Molmo2-ER           & 44.59 & 36.29 & 42.37 & 0.00 & 0.00 \\
& LLaVA-OneVision-7B & 26.96 & 13.71 & 44.97 & 0.00 & 0.00 \\
\midrule

Temp. Rec.
& Gemma-4-12B        & 96.55 & 93.10 & 41.38 & 20.69 & 20.69 \\
& Qwen3-VL-8B        & 82.76 & 82.76 & 27.59 & 3.45 & 3.45 \\
& InternVL3.5-8B     & 100.00 & 13.79 & 0.00 & 0.00 & 0.00 \\
& Molmo2-ER           & 62.07 & 44.83 & 10.34 & 0.00 & 0.00 \\
& LLaVA-OneVision-7B & 79.31 & 17.24 & 3.45 & 0.00 & 0.00 \\
\midrule

Temp. Rel.
& Gemma-4-12B        & 49.09 & 46.35 & 47.37 & 9.82 & 3.65 \\
& Qwen3-VL-8B        & 34.02 & 15.75 & 45.83 & 8.90 & 1.14 \\
& InternVL3.5-8B     & 28.54 & 23.97 & 0.06 & 0.00 & 0.00 \\
& Molmo2-ER           & 24.20 & 4.79 & 0.23 & 0.00 & 0.00 \\
& LLaVA-OneVision-7B & 29.22 & 0.23 & 0.00 & 0.00 & 0.00 \\

\bottomrule
\end{tabular}
\end{table}

\textbf{Grounding and evidence-conditioned answering expose distinct failure modes.}
Qwen reaches 62.08\% Joint@.50 and All-G@.50 of 47.42\%, indicating strong target- and question-level grounding. However, this grounding performance does not translate directly
into matched evidence-conditioned behavior with low Pair Success and JAGS@.50 of 7.49\%. In contrast, Gemma combines lower
All-G@.50 (43.18\%) with substantially stronger matched-pair behavior
(44.35\%), yielding the highest zero-shot JAGS@.50 of 21.55\%. The remaining models show that target-level presence detection alone is
insufficient for complete, question-level grounding and answering since they couple high presence accuracies between
75.59\% and 83.97\% with near-zero All-G@.50 and
JAGS@.50.

\textbf{Family-level behavior.}
Table~\ref{tab:zeroshot_perfamily} shows that answer
accuracy and grounding quality can vary independently across families.
Qwen3-VL-8B, for example, grounds Motion State evidence well
(All-G@.50=52.21\%) but almost never satisfies the matched answer--abstention
criterion for the same family (Pair=1.08\%), reducing JAGS@.50 to 0.18\%.
Conversely, Gemma-4-12B maintains stronger coupling between grounding and
evidence-conditioned answering, particularly for Spatial Relation, where it
achieves 64.63\% All-G@.50 and 36.87\% JAGS@.50. InternVL3.5-8B, Molmo2-ER, and LLaVA-OneVision-7B often attain
moderate-to-high Presence scores, yet their IoU@.50 and question-level
All-G@.50 are near zero for most families, further separating evidence presence from precise
localization. Temporal Recovery reveals the family-specific discrepancy between answering and grounding for Gemma: Pair Success reaches 93.10\%, whereas All-G@.50 is only 20.69\%. Hence, a model can exhibit the desired answering behavior without reliably recovering the full visual evidence set, or localize relevant evidence without using evidence appropriately to answer the questions. Taken together, the zero-shot results reveal distinct failure modes across architectures and reasoning families: locating relevant evidence, judging whether it is sufficient, and translating that judgment into answer-abstain behavior. This motivates our controlled adaptation study.

\subsection{Controlled Adaptation Results}
\begin{table}[t]
\centering
\caption{
Controlled adaptation study on the canonical six-family held-out
test set. Stages B--D progressively introduce evidence-sufficiency,
grounding, and explicit matched-pair supervision.
}
\label{tab:adaptation_main}
\small
\setlength{\tabcolsep}{5pt}
\textbf{(a) Evidence-conditioned answering}
\vspace{10pt}
\begin{tabular}{@{}llcccc@{}}
\toprule
\textbf{Backbone} &
\textbf{Stage} &
\textbf{Strong (\%) $\uparrow$} &
\textbf{Weak (\%) $\uparrow$} &
\textbf{Pair (\%) $\uparrow$} &
\textbf{UCR (\%) $\downarrow$} \\
\midrule

Gemma-4-12B & Zero-shot
& 67.31 & 69.47 & 44.35 & 30.42 \\

Gemma-4-12B & B
& 81.52 & \textbf{99.58} & 81.20 & \textbf{0.42} \\

Gemma-4-12B & C
& 82.58 & 99.36 & 82.05 & 0.64 \\

Gemma-4-12B & D
& 82.90 & 99.40 & 82.44 & 0.60 \\

\midrule

Qwen3-VL-8B & Zero-shot
& 54.81 & 49.26 & 17.42 & 50.74 \\

Qwen3-VL-8B & B
& \textbf{84.17} & 98.30 & \textbf{82.86} & 1.70 \\

Qwen3-VL-8B & C
& 82.86 & 98.52 & 81.70 & 1.48 \\

Qwen3-VL-8B & D
& 83.43 & 98.20 & 82.08 & 1.80 \\

\bottomrule
\end{tabular}

\textbf{(b) Grounding and joint success}
\vspace{10pt}
\begin{tabular}{@{}llrrr@{}}
\toprule
\textbf{Backbone} &
\textbf{Stage} &
\textbf{Joint@.50 (\%) $\uparrow$} &
\textbf{All-G@.50 (\%) $\uparrow$} &
\textbf{JAGS@.50 (\%) $\uparrow$} \\
\midrule
Gemma-4-12B & Zero-shot
& 60.43 & 43.18 & 21.55 \\

Gemma-4-12B & B
& 59.75 & 49.58 & 42.79 \\

Gemma-4-12B & C
& 84.41 &73.00 & 62.37 \\

Gemma-4-12B & D
& \textbf{84.87} & \textbf{73.96} & \textbf{63.29} \\

\midrule

Qwen3-VL-8B & Zero-shot
& 62.08 &47.42 & 7.49 \\

Qwen3-VL-8B & B

& 59.67 & 43.18 & 39.01 \\

Qwen3-VL-8B & C
& 83.10 & 72.44 & 62.26 \\

Qwen3-VL-8B & D
& 83.11 &72.72 & 62.83 \\

\bottomrule
\end{tabular}

\end{table}

\textbf{Evidence-sufficiency supervision primarily changes answer--abstain behavior.}
Table~\ref{tab:adaptation_main} shows a large increase in matched-pair
performance once models are explicitly trained to distinguish sufficient from
insufficient evidence. For Gemma-4-12B, Pair Success increases from 44.35\%
zero-shot to 81.20\% at Stage B, while UCR decreases from 30.42\% to 0.42\%.
The corresponding Qwen3-VL-8B change is from 17.42\% to 82.86\% Pair Success,
with UCR decreasing from 50.74\% to 1.70\%. Later stages change answer-side
metrics comparatively little: Gemma reaches 82.05\% and 82.44\% Pair Success
at Stages C and D, while Qwen reaches 81.70\% and 82.08\%, respectively.

\textbf{Grounding supervision produces a distinct localization gain.}
For Gemma, Joint@.50 increases from 59.75\% at Stage B to 84.41\% at Stage C,
while All-G@.50 increases from 49.58\% to 73.00\% and JAGS@.50 from 42.79\%
to 62.37\%. Qwen shows the same separation more clearly: Stage B already
achieves 82.86\% Pair Success, but only 59.67\% Joint@.50, 43.18\% All-G@.50,
and 39.01\% JAGS@.50. Adding grounding supervision at Stage C increases these
to 83.10\%, 72.44\%, and 62.26\%, respectively. Stage D changes these aggregate
grounding measures only slightly, reaching 83.11\% Joint@.50, 72.72\% All-G@.50,
and 62.83\% JAGS@.50. Thus, answer correctness, evidence sufficiency, and
complete visual grounding respond differently to the adaptation objectives. Appendix~\ref{app:adapted} presents detailed, per-family results of the adapted models.

%% file: sections/06_discussion.tex
\section{Discussion and Limitations}
DriveAnswer exposes a gap between producing a correct semantic answer and
determining whether the available observations justify that answer. Across all five zero-shot models, Pair Success is substantially lower than Strong Semantic Accuracy, showing that QA performance can remain high even when models fail to respond to missing evidence. Grounding provides a complementary diagnostic: correctly localizing relevant content does not  guarantee translation to correct answer-abstain behavior. For both Gemma and Qwen, evidence-sufficiency supervision accounts
for a large gain in matched-pair behavior, grounding supervision produces a
large localization improvement, while explicit coupling adds only
family-specific improvements. The cross-backbone comparison therefore supports the broader
conclusion that semantic answering, evidence sufficiency, and grounding are related but distinct capabilities that are best understood through joint evaluation.

\textbf{Limitations and Future Work.} DriveAnswer is currently constructed from nuScenes and therefore inherits its sensors, geographic coverage, annotation ontology, and temporal sampling. Future work will extend the benchmark to additional domains. Our adaptation study involves two VLM architectures that exhibit a similar trend, providing initial evidence that the observed improvements are not backbone-specific, but broader evaluation across model scales and architectures would strengthen this conclusion. Finally, evidence sufficiency is inherently question-dependent. Thus, the limited number of families also confines the benchmark's scope to the predefined tasks. Further work could include more diverse tasks and edge case tasks to ensure better generalization. 

\section{Conclusion}
We introduced DriveAnswer, a matched-evidence benchmark for evaluating reliable driving reasoning beyond answer correctness by jointly considering semantic accuracy, evidence sufficiency, and visual grounding. Its matched strong-weak construction tests whether models can answer when evidence is sufficient, abstain when it is not, and ground supported predictions in the relevant observations. Our results show that these capabilities can diverge substantially and benefit from being evaluated together. More broadly, reliable multimodal reasoning requires not only finding relevant evidence, but recognizing when that evidence is sufficient to justify an answer.
%

%% file: sections/appendix_final.tex
\appendix

\section{Benchmark Construction and Annotation Protocol}
\label{app:construction}

\subsection{Frozen benchmark population and split}
\label{app:frozen_population}

DriveAnswer contains 20,151 unique grounded questions drawn from
802 scenes and divided into seven reasoning families. Scene-level
splitting prevents questions derived from the same underlying scene
from appearing in more than one split. The frozen split contains
739 training scenes, 25 validation scenes, and 38 held-out test scenes.
No question identifier maps to more than one scene.

Each question is evaluated under two matched observation conditions.
The strong configuration preserves the observations required to justify
the verified semantic reference answer. The weak configuration changes
only the available evidence while preserving the underlying scene fact,
queried entities, and question; its intended behavior is therefore
\textsc{Abstain}. Across the frozen manifest this yields 40,302
condition instances for 20,151 unique questions.

The seven question families contain 6,098 spatial-relation questions,
1,451 map-relation questions, 6,676 motion-state questions, 2,354
temporal-relation questions, 380 temporal-recovery questions, 2,060
cross-view-recovery questions, and 1,132 evidence-conditioned decision
questions.

\subsection{End-to-End Construction Flow}
\label{app:construction_flow}

Each DriveAnswer example is constructed through the same sequence:
(1) identify an eligible verified world-state fact from the
query-independent scene representation; (2) generate a question without
exposing its reference answer; (3) derive the semantic reference independently
from privileged construction records; (4) specify the question-specific
mandatory evidence units; (5) identify and verify the observations and
grounding associated with those units; (6) assign evidence-unit support states
from the admitted observations; (7) aggregate those states into
configuration-level sufficiency; (8) construct matched strong and weak
observation configurations while preserving the underlying scene fact and
question; and (9) apply automatic integrity checks and targeted human review
before the example is admitted to the frozen benchmark.

The remainder of this section describes these stages in detail. The evaluated
model receives only the question, original unboxed images, and camera--time
identifiers; all construction annotations remain privileged.

\subsection{Question and Evidence Annotation Protocol}
\label{app:expert_protocol}
We next detail construction flow as summarized above: how questions are generated,
semantic references are recovered independently, required evidence is specified,
grounding is verified, and observation-level support is converted into
configuration-level sufficiency.
DriveAnswer separates semantic truth, visual evidence, grounding, and
sufficiency rather than deriving all labels from a single annotation step.
The protocol is organized around five logical construction components:
question generation, reference-answer recovery, required-evidence
specification, observation-support assessment, and verification. These roles
may be implemented within the same construction pipeline, but information is
kept separated so that one stage cannot determine all downstream labels.

\subsubsection{Separated Construction Roles}

\paragraph{Question generation.}
Questions are generated only from eligible verified scene records. The
semantic reference answer is not exposed to the language-based question
generator. Qwen3-VL-8B is used only to verbalize benchmark questions and referring
expressions from precomputed construction records. The prompt provides the
reasoning family, target entities, admissible descriptive attributes, and family-specific
generation constraints. It instructs the model to preserve the supplied scene
fact, avoid introducing unsupported information, use an unambiguous referring
expression, and follow the required question format. Few-shot demonstrations provide a structured input containing the reasoning family, target entity information, and admissible descriptive attributes, followed by an example natural-language question satisfying the family-specific format. The demonstrations are used only to illustrate the
desired wording and output structure; they do not provide benchmark labels for
the generated examples. Generated referring expressions are treated as candidates rather than semantic
ground truth. They are retained only when the intended target is visually
identifiable, the description is unique in the admitted observation,
same-category distractors are accounted for, and the wording does not leak the
answer. Unsupported or unstable descriptions are rewritten or rejected.
Reference answers, grounding targets, evidence requirements, and strong-weak
sufficiency labels are derived independently from the benchmark construction
records.

\paragraph{Reference-answer recovery.}
The semantic answer is recovered independently from privileged construction
information. For the six scene-understanding families, the reference is
derived from verified Layer-2 world-state facts or family-specific recovery
records. For Evidence-Conditioned Decision, the reference action is produced
by the frozen retrospective decision policy described in
Appendix~\ref{app:decision}. The reference answer is therefore fixed before
strong and weak observation configurations are compared.

\subsubsection{Family-specific evidence requirements}
\label{app:evidence_definitions}

Table~\ref{tab:family-evidence-definitions} makes explicit the distinction
between privileged world truth and the visual evidence that supports a
model-facing answer. Layer 2 facts determine the semantic reference
answer, whereas Layer 3 observations determine where that fact may be
visually supported. Layer~4 binds these components to a question-specific
evidence requirement and answer-or-abstain target.

\begin{table}[!ht]
\centering
\caption{
Question-family evidence definitions.
The strong condition preserves observations required to support
the verified semantic answer. The weak condition preserves the
underlying world-state fact while withholding the visual evidence
needed to justify that answer.
}
\label{tab:family-evidence-definitions}
\begin{tabularx}{\textwidth}{p{1.9cm}XXX}
\toprule
\textbf{Family} &
\textbf{World-state quantity} &
\textbf{Evidence required in strong} &
\textbf{Matched weak intervention} \\
\midrule

Spatial relation &
Relative spatial configuration of queried entities &
A camera observation in which the relevant entities can be identified
and their relationship established &
Remove or make insufficient the observation that establishes the
queried relationship \\

Map relation &
Relation between a queried entity and mapped scene structure &
Visual support connecting the queried entity to the relevant lane,
roadway, crossing, or other map structure &
Remove the support required to establish the entity--map relation \\

Motion state &
Whether an object is moving or stationary &
Multiple temporally separated observations preserving object identity &
Retain an anchor observation but remove the temporal evidence needed
to establish motion \\

\parbox{1.9cm}{Temporal\\relation} &
How the relation between two entities changes over time &
Temporally separated observations supporting the relational change &
Remove the observations required to establish that change \\

Temporal recovery &
Whether missing evidence at the anchor can be recovered at a nearby time &
A nearby observation of the same referent that restores the missing
evidence &
Retain only the insufficient anchor observation \\

\parbox{1.9cm}{Cross-view \\ recovery} &
Whether another synchronized view restores missing evidence &
A synchronized camera in which the same referent or relation becomes
identifiable &
Remove the supporting synchronized view while retaining the anchor \\

\parbox{1.9cm}{Evidence-conditioned\\decision} &
Reference action under the frozen conservative temporal decision policy &
Multi-camera temporal evidence sufficient to establish relevant actor
behavior and scene evolution &
Retain the synchronized anchor scene while removing the temporal
evolution required by the reference policy \\
\bottomrule
\end{tabularx}
\end{table}

\subsubsection{Grounding verification}
Grounding is verified before evidence support is assigned. Each referential
evidence unit is linked to the observation in which its visual support is
expected and, where applicable, to the corresponding spatial region.
An invalid grounding case is repaired or rejected rather than relabeled
as insufficient evidence. This distinction prevents annotation or projection
errors from being counted as legitimate weak-evidence examples.

For questions requiring multiple evidence targets, each target is verified
individually. Question-level sufficiency is not inferred from successful
grounding of only one target when multiple mandatory units are required.

\subsubsection{Observation-support assessment}
Support is assessed only from the observations admitted by the model-facing
configuration. Reviewers inspect the same unboxed observations available to
the evaluated model. Privileged world truth may be consulted to verify
identity, chronology, or a construction error, but it cannot substitute for
visual evidence when deciding whether the admitted observations support the
answer.

\subsubsection{Layer 4 Support Record}
\label{app:output_schema}

Layer~4 binds the semantic question to its required evidence, grounding
verification, support state, and model-facing target behavior. A representative
evidence-unit record for illustration is shown below:
\begin{verbatim}
{
  "question_id": "...",
  "configuration_id": "...",
  "evidence_unit_id": "...",
  "grounding_status": "VALID | BAD_GROUNDING",
  "support_state":
      "SUPPORTED | PARTIAL | AMBIGUOUS | UNAVAILABLE",
  "supporting_camera": "CAM_FRONT | ... | null",
  "supporting_sample": "... | null",
  "evidence_mechanism":
      "DIRECT | CROSS_VIEW | TEMPORAL | NONE",
  "confidence": "<value in [0,1]>",
}
\end{verbatim}
Identifiers, camera names, observation references, temporal ordering, and label
vocabularies are validated automatically. A referenced image must correspond
to the camera and sample recorded in the support record. Malformed records,
unresolved identifiers, invalid temporal links, and invalid grounding are
treated as construction failures rather than silently converted into
insufficient evidence. Table~\ref{tab:evidence_support_state} summarizes the evidence-unit support states. The \text{evidence\_mechanism} field records how support is obtained.
\text{DIRECT} indicates that the required evidence is available in the
designated observation; \text{CROSS\_VIEW} indicates recovery through a
synchronized alternative camera; \text{TEMPORAL} indicates recovery from a
different admitted time step; and \text{NONE} indicates that no admitted
observation establishes the required evidence. The fields illustrate the information retained for audit and construction; evaluated models do not receive these records.

\subsection{Support and Sufficiency Labels}
\label{app:sufficiency_rules}

Support is assigned first at the evidence-unit level and then aggregated into a
configuration-level sufficiency label. This distinction is important because a
single question may require multiple pieces of evidence.

\begin{table}[t]
\centering
\caption{Evidence-unit support states used by DriveAnswer.}
\label{tab:evidence_support_state}
\begin{tabular}{@{}lp{9.3cm}@{}}
\toprule
State & Interpretation \\
\midrule
\textsc{Supported} &
The admitted observations clearly establish the required evidence unit and its
grounding is valid. \\

\textsc{Partial} &
Relevant visual evidence is present, but an essential component needed to
establish the required fact is missing. \\

\textsc{Ambiguous} &
Relevant evidence is present, but the admitted observations permit more than
one reasonable interpretation. \\

\textsc{Unavailable} &
No admitted observation establishes the required evidence unit. \\
\bottomrule
\end{tabular}
\end{table}

\subsubsection{From Evidence Units to Configuration Sufficiency}
\label{app:sufficiency_definition}
Let $R_q$ denote the mandatory evidence units for question $q$. A configuration
is labeled \textsc{Sufficient} only when every mandatory evidence unit is
\textsc{Supported}:
\[
\mathrm{Sufficient}(q,c)
=
\mathrm{1}
\left[
\forall e\in R_q,\;
\mathrm{support}(e,c)=\textsc{Supported}
\right].
\]

If any mandatory unit is \textsc{Partial}, \textsc{Ambiguous}, or
\textsc{Unavailable}, the observations do not justify producing the semantic
answer. The model-facing target is therefore
\[
y_{q,c} =
\begin{cases}
z_q^\star, & \text{if } c \text{ is sufficient},\\
\texttt{ABSTAIN}, & \text{otherwise}.
\end{cases}
\]
Internally, configuration records may preserve the reason for insufficiency
(e.g., ambiguous versus unavailable) for auditing. At evaluation time, all
non-sufficient configurations map to the same required behavioral target,
\text{Abstain}. 

\begin{figure*}[!t]
    \centering

    \begin{minipage}[t]{0.48\linewidth}
        \centering
        \includegraphics[width=\linewidth]{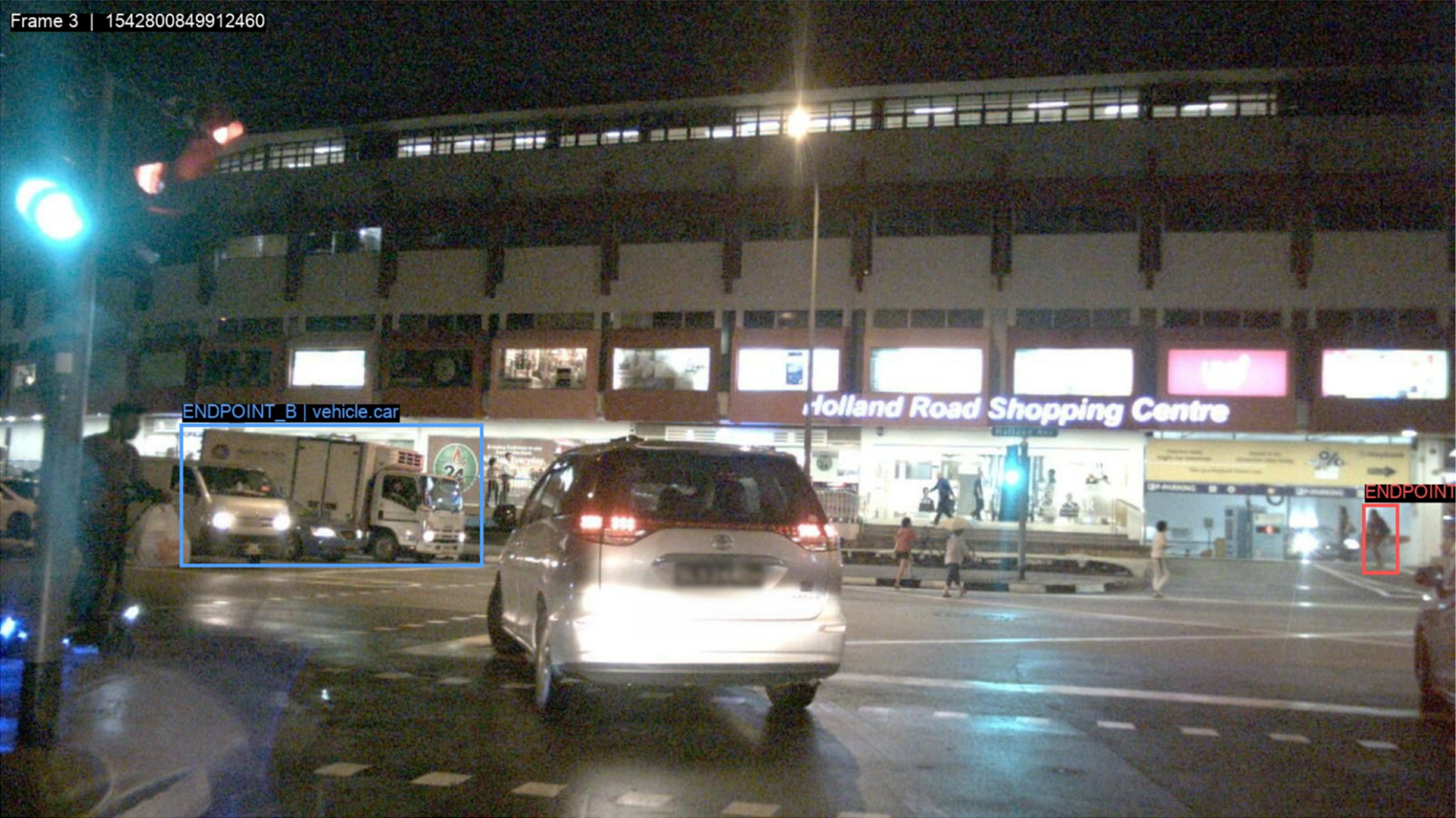}

        \small
        (a) Strong configuration contains the observations required to justify the answer.
    \end{minipage}
    \hfill
    \begin{minipage}[t]{0.48\linewidth}
        \centering
        \includegraphics[width=\linewidth]{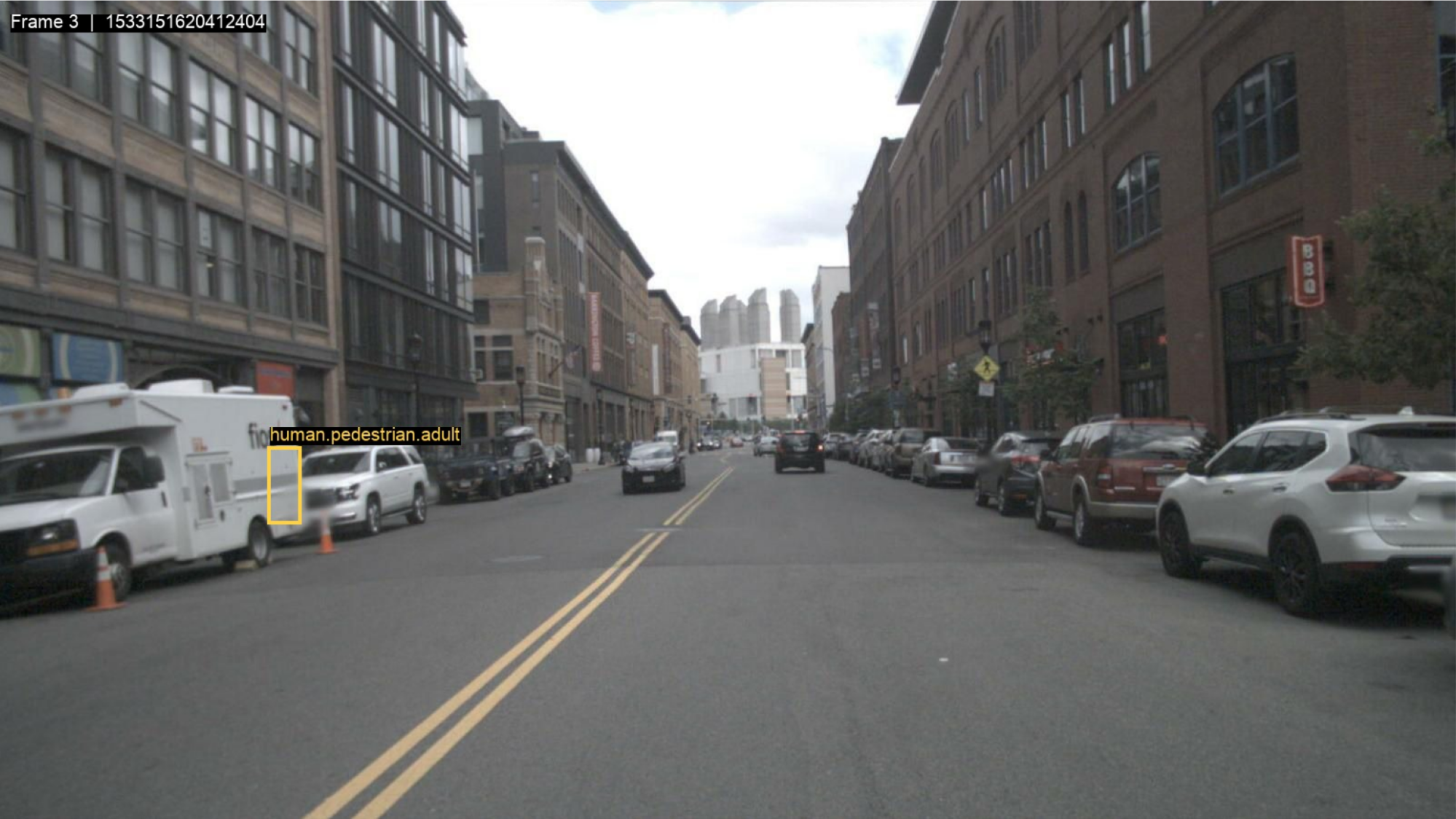}

        \small
        (b) Weak configuration does not contain enough observations required to justify the answer; thus, \textsc{Abstain} is preferred.
    \end{minipage}

    \caption{World-truth visibility does not guarantee scene observability. An actor may have a high visibility ratio in the metadata yet remain unobservable in a scene.}
    \label{fig:driveanswer_overview}
\end{figure*}

\subsection{Strong and Weak Configuration Construction}

For each retained question, DriveAnswer constructs a matched pair
$(c^{+},c^{-})$. The strong configuration $c^{+}$ must preserve every
mandatory evidence unit. The weak configuration $c^{-}$ preserves the same
question and underlying semantic fact while removing or making insufficient at
least one mandatory evidence unit (see Figure~\ref{fig:driveanswer_overview}).

Critically, weak examples are not defined simply by hiding the queried object.
For many questions, the referent remains visible in both configurations.
Instead, the intervention may remove temporal history, relational context,
map support, a synchronized alternative view, or another piece of evidence
required to justify the answer (see Figure~\ref{fig:driveanswer_overview}).

For recovery families, the anchor observation is required to be insufficient.
A cross-view or temporal recovery example is accepted only if the additional
admitted observation restores evidence that is genuinely absent from the
anchor. Cases in which the anchor already determines the answer are not
counted as recovery examples.

\subsubsection{Matched-Evidence Verification}
\label{app:matched_verification}

A candidate matched pair is retained only if all applicable checks pass:

\begin{enumerate}
    \item \textbf{World-state invariance.}
    The semantic reference answer is unchanged across strong and weak
    configurations.

    \item \textbf{Question invariance.}
    The semantic question and queried entities are identical across the pair.

    \item \textbf{Identity consistency.}
    Persistent object identity is preserved across cameras and time steps used
    by the evidence record.

    \item \textbf{Observation validity.}
    Every referenced observation resolves to an existing image, and the
    recorded camera identity and sample correspond to the actual image path.

    \item \textbf{Temporal consistency.}
    Temporal evidence is chronologically ordered and belongs to the intended
    scene and track.

    \item \textbf{Strong sufficiency.}
    Every mandatory evidence unit is supported in the strong configuration.

    \item \textbf{Weak insufficiency.}
    At least one mandatory evidence unit becomes partial, ambiguous, or
    unavailable in the weak configuration.

    \item \textbf{Criticality of removed evidence.}
    Removing the designated critical evidence changes configuration
    sufficiency.

    \item \textbf{Non-critical-context control.}
    Removing context that is not part of the mandatory evidence set must not,
    by itself, change the semantic reference answer or create an artificial
    sufficiency change.

    \item \textbf{Recovery validity.}
    For cross-view and temporal recovery, the anchor must lack the required
    evidence and the admitted synchronized or temporal observation must restore
    it.

    \item \textbf{Grounding validity.}
    Evidence loss caused by an annotation, projection, identity, or box error
    is repaired or rejected rather than treated as a valid weak intervention.
\end{enumerate}

These checks ensure that the strong--weak intervention changes the
\emph{availability of question-specific evidence} rather than changing the
underlying scene fact, question, or referent.

\paragraph{Matched pair integrity.}
\label{app:pair_integrity}
The frozen construction enforces question-level invariance across
strong and weak conditions. The semantic reference answer is derived
from the same underlying world state in both configurations, queried
entity identity is unchanged, and temporal observations remain
chronologically consistent. The intervention therefore acts on evidence
availability rather than on the underlying semantic fact.

Across the original six reasoning families, 19,019 questions contain
the target-bearing Layer 2/Layer 3 records used by the reconstruction
audit. The 1,132 evidence-conditioned decision questions use a separate
policy-relevant-actor grounding construction and are therefore audited
separately in Appendix~\ref{app:decision}.

\subsection{Automatic Validation and Review Routing}
\label{app:annotation_qc}

Construction combines deterministic checks with targeted human review.
Automatic validation verifies identifier resolution, scene membership,
camera consistency, temporal ordering, duplicate identifiers, grounding
references, matched-question identity, reference-answer invariance, and
expected strong/weak targets.

Cases are routed for additional review when the evidence state is ambiguous,
confidence is low, grounding cannot be verified automatically, or the
family-specific construction rule identifies a difficult case. Human review
assesses sufficiency from the same unboxed model-facing observations rather than
from privileged scene annotations. Privileged information is used only to
diagnose construction errors or verify identity.

Ambiguous reviewed cases are retained only when a defensible support judgment
can be established under the benchmark definition. Otherwise, they are removed
from the benchmark rather than forced into a supported or unsupported label.

\subsection{Human Validation}
\label{app:human_validation_protocol}

We perform an additional stratified validation of 120 questions from the six
original scene-understanding families, with 20 questions sampled from each
family. For each configuration, the reviewer asked whether the question could be answered confidently from the admitted observations and whether the supporting evidence could be explained in the reviewer’s own words. If the reviewer could not confidently determine and justify the answer from the available observations, the configuration was not judged sufficient. The reviewer assesses four properties independently: whether the strong
configuration contains sufficient evidence, whether the weak configuration is
insufficient, whether the queried referent is clear, and whether the example
admits an unintended alternative interpretation.

The reviewer agrees that the strong configuration is sufficient for
117/120 examples (97.5\%) and that the weak configuration is insufficient for
117/120 examples (97.5\%). Referent clarity is accepted for 111/120 examples
(92.5\%), and absence of unintended ambiguity for 115/120 examples (95.8\%).
These are marginal criterion counts and are not interpreted as a 97.5\%
matched-pair acceptance rate.

The Decision family is reviewed separately because its semantic reference is
policy-conditioned rather than a direct scene fact. A stratified set of
60 held-out Decision questions contains all 10 \textsc{Proceed} examples and
25 each from \textsc{Slow} and \textsc{Stop}. The frozen reference action
agrees with manual assessment in 57/60 cases (95.0\%): 10/10 for
\textsc{Proceed}, 22/25 for \textsc{Slow}, and 25/25 for \textsc{Stop}.
These reviews are reported as construction-quality diagnostics rather than as
model-performance measurements as in Table~\ref{tab:human-review}.

\begin{table}[!ht]
\centering
\caption{Human-review diagnostics across the original six families and
the Decision-family label audit.}
\label{tab:human-review}
\begin{tabular}{lr}
\toprule
\textbf{Criterion} & \textbf{Agreement} \\
\midrule
Strong condition sufficient & 117/120 (97.5\%) \\
Weak condition insufficient & 117/120 (97.5\%) \\
Referent clear & 111/120 (92.5\%) \\
No unintended ambiguity & 115/120 (95.8\%) \\
\midrule
Decision reference action, overall & 57/60 (95.0\%) \\
\quad PROCEED & 10/10 (100.0\%) \\
\quad SLOW & 22/25 (88.0\%) \\
\quad STOP & 25/25 (100.0\%) \\
\bottomrule
\end{tabular}
\end{table}

\section{Graph and Evidence Validation}
\label{app:graph_validation}

\subsection{Graph construction and reconstruction audits}
\label{app:graph-audit}

The final graph collection contains 34,149 scene graphs with
7,875,985 nodes and 89,533,730 typed edges. Automated integrity checks
found no broken structural references. Geometric projection succeeds
for 99.8361\% of eligible projection attempts. Temporal construction
uses an exact five-frame neighborhood centered on the anchor observation
when available.

We additionally reconstruct the privileged fact associated with each
target-bearing question independently from its observation-level support
record. Among 19,019 target-bearing questions, 18,960 reconstruct
directly, corresponding to 99.690\%. The remaining 59 normalization
exceptions occur in the spatial-relation family and are retained as
explicit normalization exceptions rather than silently altering the
underlying relation labels. The construction audit verifies that every retained strong configuration contains support for all mandatory evidence units and that every matched weak configuration removes support for at least one mandatory unit. These 100\% rates therefore measure compliance with the frozen construction rules rather than independent perceptual agreement. The support rates verify that the benchmark satisfies its strong-weak construction rules; they are not independent estimates of human perceptual sufficiency, which is evaluated separately in Appendix~\ref{app:human_validation_protocol}. Tables~\ref{tab:graph-audit}-\ref{tab:observation-support-audit} summarize the graph reconstruction, matched-pair integrity, and observation-support audits.

\begin{table}[!ht]
\centering

\caption{Full graph construction audit.
}
\label{tab:graph-audit}
\begin{tabular}{lr}
\toprule
\textbf{Audit quantity} & \textbf{Result} \\
\midrule    
Scene graphs & 34,149 \\
Graph nodes & 7,875,985 \\
Graph edges & 89,533,730 \\
Structural integrity errors & 0 \\
Projection success & 99.8361\% \\
Strong Layer 2 answer preserved & 100\% \\
Weak \text{Abstain} target & 100\% \\
Temporal neighborhood & 5 frames \\
\bottomrule
\end{tabular}
\end{table}

\begin{table}
\centering
\captionof{table}{Layer 2 to Layer 3 reconstruction audit.
}
\label{tab:l2-l3-audit}

\begin{tabular}{lr}
\toprule
\textbf{Quantity} & \textbf{Result} \\
\midrule
Target-bearing questions & 19,019 \\
Directly reconstructed & 18,960 \\
Reconstruction rate & 99.690\% \\
Normalization exceptions & 59 \\
\bottomrule
\end{tabular}

\end{table}

\subsection{Observation-support audit}
\label{app:visibility_audit}
Projection of an annotated target into a camera is not treated as proof
of perceptual sufficiency. The benchmark separately records
observation-level support and question-specific evidence requirements.
All strong configurations preserve the family-specific support required
by construction under the benchmark construction protocol, whereas the matched weak configuration removes that
required support.

For the 19,019 target-bearing questions, all required targets are
visually present under the strong configuration. Under weak
configurations, 60.308\% retain all target objects as visible while
39.692\% make at least one target unavailable. This distinction is
important: a weak configuration may leave the queried objects visibly
present while removing the temporal, cross-view, map, or relational
evidence necessary to justify the answer.

\begin{table}[!ht]
\centering
\caption{
Matched-pair integrity audit.
}
\label{tab:pair-integrity-audit}
{
\begin{tabular}{lr}
\toprule
\textbf{Check} & \textbf{Pass rate} \\
\midrule
Strong semantic target preserved & 100.0\% \\
Weak target requires abstention & 100.0\% \\
Strong reference matches world truth & 100.0\% \\
Weak reference matches abstention target & 100.0\% \\
Strong/weak observations resolve & 100.0\% \\
\bottomrule
\end{tabular}
}
\end{table}

\begin{table}[!ht]
\centering
\caption{Construction-level target visibility and required-evidence audit.
}
\label{tab:observation-support-audit}

{
\begin{tabular}{lr}
\toprule
\textbf{Quantity} & \textbf{Result} \\
\midrule
Target-bearing questions & 19,019 \\
Strong: targets visible & 100.000\% \\
Weak: all targets visible & 60.308\% \\
Weak: target unavailable & 39.692\% \\
Strong: required support present by construction & 100.000\% \\
Weak: $\geq1$ mandatory support unit removed by construction & 100.000\% \\
\bottomrule
\end{tabular}%
}

\end{table}

\subsection{Visibility Triage Rather Than Visibility Truth}
\label{app:visibility-risk}

The visibility procedure was enforced because geometric projection and source visibility metadata are useful but incomplete. The augmentation therefore combines several cues such as target size, boundary truncation, source visibility, estimated occlusion, foreground overlap/crowding, and blur when measurable into a review-routing record. We freeze the routing thresholds only after calibration on labeled cases. The exact score is not released as a truth label because two observations with similar risk can differ in whether the evidence needed by a particular question is actually identifiable. This distinction is notable for occlusion. Partial occlusion does not automatically imply abstention. A reviewer or support expert instead asks whether the required entity or relation can be established from the admitted observations. The triage stage only decides which cases deserve closer inspection.

\subsection{Visibility-only diagnostic}
\label{app:visibility_diagnostic}

We present the visibility-only diagnostic as an
oracle-support sanity check, not as a learned baseline as detailed in Table~\ref{tab:visibility-diagnostic}. The diagnostic
uses privileged construction annotations and therefore must not be
interpreted as ordinary model performance. A permissive target-visibility
rule obtains 37.21\% matched-pair success, while a stricter observability
rule reaches 43.89\%. These results show that target visibility is informative but insufficient for determining question-specific evidence sufficiency: even privileged visibility rules fail on more than half of the matched pairs. Thus, visibility alone cannot substitute for reasoning over the temporal, cross-view, relational, or map evidence required by the question.

\begin{table}[!ht]
\centering
\caption{
Visibility diagnostic.
These oracle-support rules use privileged observation annotations and
are not model baselines.
}
\label{tab:visibility-diagnostic}
\begin{tabular}{lrrrrr}
\toprule
\textbf{Rule} & \textbf{Strong (\%) $\uparrow$} & \textbf{Weak (\%) $\uparrow$} & \textbf{Pair (\%) $\uparrow$} & \textbf{UCR (\%) $\downarrow$} & \textbf{UAR (\%) $\downarrow$} \\
\midrule
Permissive visibility
& 100.00 & 37.21 & 37.21 & 62.79 & 0.00 \\
Strict observability
& 98.55 & 45.34 & 43.89 & 54.66 & 1.45 \\
\bottomrule
\end{tabular}
\end{table}

\section{Evidence-Conditioned Decision Family}
\label{app:decision}

\subsection{Decision construction}
\label{app:decision_construction}

The evidence-conditioned decision family contains 1,132 questions:
869 training, 102 validation, and 161 held-out test questions. Decision labels are not treated as direct physical world-state facts.
Instead, \text{Proceed}, \text{Slow}, and \text{Stop} are produced
via human-designed rules.
This policy operates on verified actor trajectories and temporal conflict
signals and uses fixed construction thresholds: an 8\,m stop distance,
20\,m slow distance, 12\,m close-caution distance, 25\,m clearance
threshold, 15\,m separating-clearance threshold, and a 1\,m
relative-change threshold. The presence of leading and following and applicable traffic rules is also accounted for. Complicated and ambiguous construction cases such as driving in pedestrian-dense areas or driving near emergency vehicles are flagged for manual reviews. Because each example contains observations on both sides of the
anchor, this family evaluates \emph{retrospective temporal-window
reasoning}. It must not be interpreted as online closed-loop driving or
as a policy that acts before future observations become available.

\subsection{Decision distribution and evidence statistics}

\begin{table}[!ht]
\centering

\caption{Decision-family label distribution.
}
\label{tab:decision-labels}
{
\begin{tabular}{lrrrr}
\toprule
\textbf{Split} & \textbf{Proceed} & \textbf{Slow} & \textbf{Stop} & \textbf{Total} \\
\midrule
Train & 61 & 690 & 118 & 869 \\
Validation & 0 & 96 & 6 & 102 \\
Test & 10 & 124 & 27 & 161 \\
\bottomrule
\end{tabular}%
}
\end{table}

\begin{table}
\centering
\caption{Decision-family evidence statistics.
}
\label{tab:decision-evidence}
{
\begin{tabular}{lr}
\toprule
\textbf{Quantity} & \textbf{Result} \\
\midrule
Mean relevant actors, train & 3.474 \\
Mean relevant actors, validation & 2.441 \\
Mean relevant actors, test & 3.236 \\
Approaching actor trends & 2,095 \\
Stable actor trends & 1,463 \\
Separating actor trends & 231 \\
Proceed examples with no relevant actor & 60 \\
\bottomrule
\end{tabular}
}

\end{table}

The 161 Decision test questions contain 521 unique policy-relevant actor
targets, an average of 3.236 actors per question. Further statistics are reported in Tables~\ref{tab:decision-labels} and~\ref{tab:decision-evidence}. The number ranges from
zero to 18. Eight test questions contain no policy-relevant actor and
153 contain at least one. For grounding evaluation, actor-observation localization instances are
enumerated separately from unique policy-relevant actor identities. The
Decision grounding evaluator therefore contains 1,058 target instances
over the same 161 test questions. We keep these denominators separate:
521 describes unique actors used by the reference policy, whereas 1,058
counts actor-observation grounding targets.

\subsection{Decision shortcut controls}
\label{app:decision_shortcuts}

The held-out Decision label distribution is imbalanced:
124 \textsc{Slow}, 27 \textsc{Stop}, and 10 \textsc{Proceed}.
An always-\textsc{Slow} predictor reaches 77.02\% accuracy, but only
29.01\% macro-F1 and 33.33\% balanced accuracy. Indicating monotonic policy are not suffice for determining the correct driving decision. We additionally evaluate a count-only baseline using the number of policy-relevant actors. For every
actor count observed during training, the baseline predicts the majority
training label associated with that count. The learned rule maps zero
actors to \textsc{Proceed} and every observed positive actor count to
\textsc{Slow}. The count-only model obtains 81.99\% accuracy, 59.47\% macro-F1, and
60.00\% balanced accuracy (see Table~\ref{tab:decision-shortcuts}). This association is partly expected: clear scenes with no relevant actors tend to support \textsc{Proceed}, while scenes with relevant actors require the ego vehicle to act more cautiously. More severe cases requiring \textsc{Stop} are correspondingly rarer. However, actor count alone does not determine the action: the count-only baseline
predicts none of the 27 \textsc{Stop} questions correctly. The number of actors is treated as privileged construction information that is not provided to the model at evaluation; thus this experiment is a distributional diagnostic and not a model-facing baseline.

\begin{table}[!ht]
\centering

\caption{Decision-family shortcut baselines.
}
\label{tab:decision-shortcuts}
{
\begin{tabular}{lrrr}
\toprule
\textbf{Baseline} & \textbf{Accuracy (\%) $\uparrow$} & \textbf{Macro-F1 (\%) $\uparrow$} & \textbf{Balanced accuracy (\%) $\uparrow$} \\
\midrule
Always \textsc{Slow}
& 77.02 & 29.01 & 33.33 \\
Actor-count only
& 81.99 & 59.47 & 60.00 \\
\bottomrule
\end{tabular}%
}
\end{table}
\section{Evaluation and Statistical Reliability}
\label{app:statistics}
\subsection{Evaluation metrics}

For each matched question pair, Strong Semantic Accuracy measures
whether the strong-condition answer equals the semantic reference.
Weak Abstention Accuracy measures whether the weak-condition response is
\text{Abstain}. Matched-Pair Success requires both events for the same
question. Unsupported Commitment Rate (UCR) measures parsed
non-\text{Abstain} commitments under weak evidence. Unnecessary Abstention Rate (UAR)
measures abstention on answerable strong inputs.

\subsection{All-G and JAGS}

For a question with multiple required grounding targets, All-G@$t$ equals one
iff every required target achieves Joint@$t$ success. Joint Answer--Grounding
Success is then defined as
\[
\mathrm{JAGS}@t
=
\mathrm{1}\left[
\mathrm{PairSuccess}
\land
\mathrm{AllG}@t
\right].
\]
Because answering and grounding are elicited through separate calls,
JAGS measures co-occurring answer, abstention, and localization success.
It is not interpreted as evidence that the predicted boxes were the
causal internal explanation of the semantic answer.

\subsection{Scene-clustered bootstrap}
\label{app:bootstrap}

\subsubsection{6-family Bootstrap Results}
\label{app:6fam_bootstrap}

We estimate uncertainty with a scene-clustered bootstrap over the
six-family test set. We choose scene-clustered bootstrapping because matched strong-weak questions and grounding targets may originate from the same scene and hence are not statistically independent. Moreover, metrics such as All-G and JAGS are question-level conjunctions, which requires bootstrapping of scenes rather than parametric variance tests. We therefore estimate uncertainty by resampling held-out scenes rather than individual questions or targets. The six-family scene-clustered bootstrap confirms that the main adaptation
patterns are not driven by a small number of held-out scenes (see Table~\ref{tab:sixfamily-bootstrap}). For Gemma,
Stage~C reaches 82.58\% Strong Accuracy
(95\% CI: [78.80, 86.16]), 82.05\% Pair Success
([78.14, 85.70]), and 62.37\% JAGS@.50
([55.43, 69.39]); Stage~D is similar at 82.90\%
([79.26, 86.29]), 82.44\% ([78.66, 85.93]), and
63.29\% ([56.41, 70.23]), respectively.
Qwen shows the same stability between the final two stages:
Stage~C obtains 82.86\% Strong Accuracy
([78.94, 86.62]), 81.70\% Pair Success
([77.63, 85.59]), and 62.26\% JAGS@.50
([54.62, 69.94]), compared with 83.43\%
([79.39, 87.16]), 82.08\% ([77.77, 86.05]), and
62.83\% ([55.06, 70.50]) at Stage~D.
The corresponding grounding intervals likewise overlap strongly:
Qwen Joint@.50 is 83.10\% [78.72, 87.43] at Stage~C and
83.11\% [78.76, 87.49] at Stage~D.

\begin{table}[t]
\centering
\caption{
Selected results on the canonical six-family held-out test set with
95\% scene-clustered bootstrap intervals. All systems are evaluated on
the same 2,830 matched questions. $t=0.50$.
}
\label{tab:sixfamily-bootstrap}
{
\begin{tabular}{lccc}
\toprule
\textbf{System}
& \textbf{Strong (\%) $\uparrow$}
& \textbf{Pair (\%) $\uparrow$}
& \textbf{JAGS@$t$ (\%) $\uparrow$} \\
\midrule
Gemma-C
& 82.58 [78.80, 86.16]
& 82.05 [78.14, 85.70]
& 62.37 [55.43, 69.39] \\

Gemma-D
& 82.90 [79.26, 86.29]
& 82.44 [78.66, 85.93]
& 63.29 [56.41, 70.23] \\

Qwen-C
& 82.86 [78.94, 86.62]
& 81.70 [77.63, 85.59]
& 62.26 [54.62, 69.94] \\

Qwen-D
& 83.43 [79.39, 87.16]
& 82.08 [77.77, 86.05]
& 62.83 [55.06, 70.50] \\
\bottomrule
\end{tabular}
}
\end{table}

\begin{table}[!ht]
\centering
\caption{
Stage C minus Stage D paired differences on the canonical six-family test set.
Values are percentage points with 95\% scene-clustered intervals.
}
\label{tab:paired-cd}
\begin{tabular}{lrr}
\toprule
\textbf{Metric} & \textbf{Gemma C--D} & \textbf{Qwen C--D} \\
\midrule
Strong (\%) $\uparrow$
& -0.32 [-1.25, +0.56]
& -0.57 [-1.84, +0.50] \\

Pair (\%) $\uparrow$
& -0.39 [-1.32, +0.50]
& -0.39 [-1.70, +0.73] \\

UCR (\%) $\downarrow$
& +0.04 [-0.15, +0.25]
& -0.32 [-0.67, +0.00] \\

Joint@.50 (\%) $\uparrow$
& -0.45 [-1.02, +0.08]
& -0.02 [-0.38, +0.34] \\

All-G@.50 (\%) $\uparrow$
& -0.95 [-2.00, +0.00]
& -0.28 [-0.89, +0.32] \\

JAGS@.50 (\%) $\uparrow$
& -0.92 [-2.09, +0.18]
& -0.57 [-1.65, +0.38] \\
\bottomrule
\end{tabular}
\end{table}

The paired analysis in Table~\ref{tab:paired-cd} provides a stricter
comparison because both stages are evaluated on exactly the same questions
and bootstrap scene draws. For Gemma, the C--D differences are
$-0.32$ points for Strong Accuracy
(95\% CI: [-1.25, +0.56]), $-0.39$ for Pair Success
([-1.32, +0.50]), $-0.45$ for Joint@.50
([-1.02, +0.08]), and $-0.92$ for JAGS@.50
([-2.09, +0.18]). Qwen is similarly close:
the C--D differences are $-0.57$ for Strong Accuracy
([-1.84, +0.50]), $-0.39$ for Pair Success
([-1.70, +0.73]), $-0.02$ for Joint@.50
([-0.38, +0.34]), and $-0.57$ for JAGS@.50
([-1.65, +0.38]). Most intervals contain zero, so the six-family results
do not support a consistent aggregate improvement from Stage~C to Stage~D.

The scene-clustered analysis reinforces the separation between answer behavior and grounding. For both backbones, Stage B$\rightarrow$C produces large gains in grounding and JAGS while changes in answer-side metrics are comparatively smaller; subsequent Stage C$\rightarrow$D changes are small.

\section{No-image+Metadata Shortcut Controls}
\label{app:shortcut_controls}

We evaluate a zero-shot Gemma-4-12B under the 2,991-pair
test protocol after removing both image placeholders and image pixels and report the results in Table~\ref{tab:no-image-gemma}.
The question text, allowed label set, and camera and time observation
metadata are left unchanged. We therefore refer to this setting as the
\emph{no-image text and metadata baseline}. Without visual input, Gemma strongly defaults to abstention:
Strong Semantic Accuracy falls to 3.31\%, Weak Abstention Accuracy is
99.36\%, Matched-Pair Success is 3.08\%, UCR is 0.47\%, and UAR reaches
96.42\%. Of 2,991 strong examples, 2,884 produce \text{Abstain}. The family breakdown is also informative. Strong accuracy is
11.94\% for cross-view recovery, 0.00\% for map relation, 1.71\% for
motion state, 4.72\% for spatial relation, 51.72\% for temporal
recovery, 0.00\% for temporal relation, and 0.00\% for Decision.
The unusually high temporal-recovery value suggests residual task or
language priors in that small family and is treated as a limitation
rather than evidence of visual reasoning. This control indicates that the benchmark as presented to a zero-shot model is not trivially solvable from question wording and camera and metadata alone. 

\begin{table}
\centering
\captionof{table}{Gemma no-image text+metadata control.
}
\label{tab:no-image-gemma}
{
\begin{tabular}{rrrrr}
\toprule
\textbf{Strong (\%) $\uparrow$} & \textbf{Weak (\%) $\uparrow$} & \textbf{Pair (\%) $\uparrow$} & \textbf{UCR (\%) $\downarrow$} & \textbf{UAR (\%) $\downarrow$} \\
\midrule
3.31 & 99.36 & 3.08 & 0.47 & 96.42 \\
\bottomrule
\end{tabular}%
}

\end{table}
\section{Zero-Shot Baselines }
\label{app:zero_shot}
\subsection{Zero-shot Answering Performance per Family}
Table~\ref{tab:zeroshot-family} reveals that aggregate zero-shot results conceal substantial differences across reasoning families. Recovery tasks are comparatively favorable for some models: Gemma reaches 93.10\% Pair Success on Temporal Recovery, while Qwen reaches 82.76\%. In contrast, Motion State illustrates a different failure mode: abstention is rare among all models, reducing Pair Success to 1.08\% for Qwen, 2.98\% for InternVL, 0.54\% for Molmo, and 0.09\% for LLaVA. 

Decision is the clearest shared zero-shot weakness. Gemma, Qwen, InternVL, and Molmo obtain zero matched-pair success, while LLaVA reaches only 0.62\%. Map, Spatial, and Temporal Relation families show intermediate and model-dependent behavior. These differences indicate that the overall zero-shot failure is not attributable to a single reasoning family.

\begin{table*}[!ht]
\small
\centering
\caption{
Per-family zero-shot answering under the seven-family held-out protocol.
Pair requires a correct strong-condition answer and correct abstention on the
matched weak condition. \textbf{Results are expressed as percentages.}
}
\label{tab:zeroshot-family}
{
\begin{tabular}{llrrrrr}
\toprule
\textbf{Family}
& \textbf{Model}
& \textbf{Strong $\uparrow$}
& \textbf{Weak $\uparrow$}
& \textbf{Pair $\uparrow$}
& \textbf{UCR $\downarrow$}
& \textbf{UAR $\downarrow$} \\
\midrule

\parbox{1.8cm}{Cross-view \\recovery}
& Gemma-4-12B
& 54.73 & 99.00 & \textbf{53.73} & 1.00 & 45.27 \\

& Qwen3-VL-8B
& 51.74 & \textbf{100.00} & 51.74 & \textbf{0.00} & 48.26 \\

& InternVL3.5-8B
& 60.70 & 45.27 & 19.90 & 54.73 & 39.30 \\

& Molmo2-ER
& 20.40 & 97.01 & 18.41 & 2.99 & 79.60 \\

& LLaVA-OneVision-7B
& \textbf{73.13} & 53.23 & 32.34 & 46.77 & \textbf{26.87} \\

\midrule

Decision
& Gemma-4-12B
& 18.63 & 0.00 & 0.00 & 100.00 & \textbf{0.00} \\

& Qwen3-VL-8B
& 13.66 & 0.00 & 0.00 & 100.00 & \textbf{0.00} \\

& InternVL3.5-8B
& \textbf{63.35} & 0.00 & 0.00 & 100.00 & 0.62 \\

& Molmo2-ER
& 6.21 & 0.00 & 0.00 & 100.00 & \textbf{0.00} \\

& LLaVA-OneVision-7B
& 34.78 & \textbf{2.48} & \textbf{0.62} & \textbf{97.52} & 1.86 \\

\midrule

Map relation
& Gemma-4-12B
& \textbf{54.59} & \textbf{95.14} & \textbf{51.89}
& \textbf{4.86} & 5.41 \\

& Qwen3-VL-8B
& 43.24 & 68.65 & 25.95 & 31.35 & 0.54 \\

& InternVL3.5-8B
& 39.46 & 31.35 & 7.03 & 68.65 & \textbf{0.00} \\

& Molmo2-ER
& 41.08 & 9.73 & 4.32 & 90.27 & \textbf{0.00} \\

& LLaVA-OneVision-7B
& 36.76 & 82.16 & 28.65 & 17.84 & 14.05 \\

\midrule

Motion state
& Gemma-4-12B
& 76.28 & \textbf{44.72} & \textbf{33.90}
& \textbf{55.28} & \textbf{0.00} \\

& Qwen3-VL-8B
& \textbf{83.23} & 1.71 & 1.08 & 98.29 & 0.09 \\

& InternVL3.5-8B
& 82.96 & 4.87 & 2.98 & 95.13 & 1.08 \\

& Molmo2-ER
& 80.25 & 1.53 & 0.54 & 98.47 & 0.09 \\

& LLaVA-OneVision-7B
& 46.62 & 0.63 & 0.09 & 99.37 & 0.45 \\

\midrule

Spatial relation
& Gemma-4-12B
& \textbf{69.70} & 73.85 & \textbf{51.27}
& 26.15 & \textbf{4.95} \\

& Qwen3-VL-8B
& 31.22 & \textbf{92.17} & 27.19
& \textbf{7.83} & 18.09 \\

& InternVL3.5-8B
& 34.56 & 73.39 & 22.24 & 26.61 & 24.19 \\

& Molmo2-ER
& 44.59 & 83.18 & 36.29 & 16.82 & 26.61 \\

& LLaVA-OneVision-7B
& 26.96 & 63.82 & 13.71 & 36.18 & 11.18 \\

\midrule

\parbox{1.8cm}{Temporal\\recovery}
& Gemma-4-12B
& 96.55 & 96.55 & \textbf{93.10}
& 3.45 & 3.45 \\

& Qwen3-VL-8B
& 82.76 & \textbf{100.00} & 82.76
& \textbf{0.00} & 17.24 \\

& InternVL3.5-8B
& \textbf{100.00} & 13.79 & 13.79
& 86.21 & \textbf{0.00} \\

& Molmo2-ER
& 62.07 & 82.76 & 44.83 & 17.24 & 37.93 \\

& LLaVA-OneVision-7B
& 79.31 & 24.14 & 17.24 & 75.86 & 20.69 \\

\midrule

\parbox{1.8cm}{Temporal \\ relation}
& Gemma-4-12B
& \textbf{49.09} & \textbf{97.26} & \textbf{46.35}
& \textbf{2.74} & 9.13 \\

& Qwen3-VL-8B
& 34.02 & 49.77 & 15.75 & 50.23 & 16.44 \\

& InternVL3.5-8B
& 28.54 & 92.01 & 23.97 & 7.99 & 16.44 \\

& Molmo2-ER
& 24.20 & 42.24 & 4.79 & 57.76 & 20.32 \\

& LLaVA-OneVision-7B
& 29.22 & 0.68 & 0.23 & 99.32 & \textbf{0.00} \\

\bottomrule
\end{tabular}%
}
\end{table*}

\subsection{Zero-shot Grounding Performance per Family}

\begin{table}[!ht]
\centering
\scriptsize
\caption{Per-family zero-shot grounding results on the held-out
test set. Models are grouped within each question family to expose
family-specific differences. Joint@.50 evaluates target-level grounding;
All-G@.50 requires every target associated with a question to be grounded;
JAGS@.50 additionally requires matched-pair QA success. We report $t=0.50$. \textbf{All values are expressed as percentages.} Cross is Cross-view Recovery; Map Rel. is Map Relation; Temp. Rec. is Temporal Recovery; Temp. Rel. is Temporal Relation. Crossview N=201; Map Relation N=185, Decision N=161; Spatial Relation N=868, Motion State N=1109, Temporal Recovery N=29, Temporal Relation N=438.
}
\label{tab:zeroshot-family-detailed}
{
\begin{tabular}{@{}p{0.8cm}p{2.3cm}rrrrrrr@{}}
\toprule
\textbf{Family} & \textbf{Model} 
& \textbf{Presence} $\uparrow$ & \textbf{IoU@$t$} $\uparrow$ & \textbf{Joint@$t$} $\uparrow$
& \textbf{All-G@$t$} $\uparrow$ & \textbf{JAGS@$t$} $\uparrow$ \\
\midrule

\multirow{5}{*}{\parbox{1.5cm}{Cross.}}
& Gemma-4-12B           & \textbf{74.38} & 19.06 & 14.18 & 2.49 & \textbf{1.99} \\
& Qwen3-VL-8B           & 58.71 & \textbf{39.83} & \textbf{23.38} & \textbf{2.99} & \textbf{1.99} \\
& InternVL3.5-8B        & 38.31 & 0.00 & 0.00 & 0.00 & 0.00 \\
& Molmo2-ER             & 73.63 & 0.34 & 0.25 & 0.00 & 0.00 \\
& LLaVA-OneVision-7B    & 56.22 & 0.00 & 0.00 & 0.00 & 0.00 \\
\midrule

\multirow{5}{*}{\parbox{1.5cm}{Map\\Rel.}}
& Gemma-4-12B           & \textbf{94.05} & 61.67 & 75.41 & 54.05 & \textbf{29.73} \\
& Qwen3-VL-8B           & 89.46 & \textbf{72.73} & \textbf{78.11} & \textbf{57.30} & 15.14 \\
& InternVL3.5-8B        & 83.51 & 0.00 & 48.65 & 0.00 & 0.00 \\
& Molmo2-ER             & 89.19 & 0.67 & 49.19 & 0.54 & 0.00 \\
& LLaVA-OneVision-7B    & 88.11 & 0.00 & 49.19 & 0.00 & 0.00 \\
\midrule

\multirow{5}{*}{\parbox{1.5cm}{Deci\\sion}}
& Gemma-4-12B          52.74 & \textbf{17.57} & 9.74 & 2.48 & 0.00\\
& Qwen3-VL-8B          & \textbf{86.48} & 13.99 & \textbf{12.67} & \textbf{14.29} & 0.00 \\
& InternVL3.5-8B       & 46.88 & 0.00 & 1.04 & 1.86 & 0.00 \\
& Molmo2-ER             & 94.90 & 0.20 & 0.76 & 0.00 & 0.00 \\
& LLaVA-OneVision-7B    & 13.99 & 0.00 & 1.51 & 4.97 & \textbf{0.62} \\
\midrule

\multirow{5}{*}{\parbox{1.5cm}{Spatial\\Rel.}}
& Gemma-4-12B           & \textbf{91.70} & 76.72 & 80.09 & 64.63 & \textbf{36.87} \\
& Qwen3-VL-8B           & 89.99 & \textbf{84.97} & \textbf{83.13} & \textbf{70.39} & 19.82 \\
& InternVL3.5-8B       & 87.56 & 0.26 & 45.63 & 0.23 & 0.12 \\
& Molmo2-ER             & 79.20 & 0.00 & 42.37 & 0.00 & 0.00 \\
& LLaVA-OneVision-7B    & 88.33 & 0.00 & 44.97 & 0.00 & 0.00 \\
\midrule

\multirow{5}{*}{\parbox{1.5cm}{Motion\\State}}
& Gemma-4-12B          & \textbf{88.64} & 68.92 & 61.09 & 45.72 & \textbf{18.85} \\
& Qwen3-VL-8B          & 85.71 & \textbf{73.49} & \textbf{62.98} & \textbf{52.21} & 0.18 \\
& InternVL3.5-8B       & 75.16 & 0.00 & 0.00 & 0.00 & 0.00 \\
& Molmo2-ER             & 85.03 & 0.21 & 0.18 & 0.00 & 0.00 \\
& LLaVA-OneVision-7B  & 75.07 & 0.00 & 0.00 & 0.00 & 0.00 \\
\midrule

\multirow{5}{*}{\parbox{1.5cm}{Temp.\\Rec.}}
& Gemma-4-12B         & 77.59 & \textbf{53.33} & \textbf{41.38} & \textbf{20.69} & \textbf{20.69} \\
& Qwen3-VL-8B           & 62.07 & 44.44 & 27.59 & 3.45 & 3.45 \\
& InternVL3.5-8B        & 48.28 & 0.00 & 0.00 & 0.00 & 0.00 \\
& Molmo2-ER           & \textbf{81.03} & 12.77 & 10.34 & 0.00 & 0.00 \\
& LLaVA-OneVision-7B    & 75.86 & 4.55 & 3.45 & 0.00 & 0.00 \\
\midrule

\multirow{5}{*}{\parbox{1.5cm}{Temp.\\Rel.}}
& Gemma-4-12B          & \textbf{89.27} & 53.07 & \textbf{47.37} & \textbf{9.82} & \textbf{3.65} \\
& Qwen3-VL-8B         & 81.85 & \textbf{56.00} & 45.83 & 8.90 & 1.14 \\
& InternVL3.5-8B       & 72.26 & 0.08 & 0.06 & 0.00 & 0.00 \\
& Molmo2-ER             & 88.93 & 0.26 & 0.23 & 0.00 & 0.00 \\
& LLaVA-OneVision-7B    & 64.90 & 0.00 & 0.00 & 0.00 & 0.00 \\
\bottomrule
\end{tabular}%
}
\end{table}

The family-level results reveal distinct grounding failure modes across reasoning types as detailed in Table~\ref{tab:zeroshot-family-detailed}. Spatial Relation and Map Relation are favorable for Gemma and Qwen, both of which achieve substantially higher target-level Joint@0.5 and question-level All-G@0.5 than the remaining zero-shot models. Temporal Recovery is also relatively successful for Gemma at the target level, although complete-evidence recovery remains considerably lower. Cross-view Recovery and Decision are substantially harder since target-level grounding does not reliably translate into complete question-level evidence recovery (Qwen reaches 23.38\% for Joint@0.5 but only 2.99\% for All-G@0.5). Decision is similarly challenging, with high presence accuracy among many models but weak localization and JAGS, indicating that detecting relevant actors is substantially easier than localizing the complete evidence needed for meaningful answering. The weaker grounding models exhibit this discrepancy among the majority of reasoning tasks. Together, the zero-shot models support the joint reporting of presence detection, target localization, complete evidence recovery, and matched evidence-answering because they simultaneously measure different aspects of models' performance.
\section{Adapted Models}
\label{app:adapted}
\subsection{Exact Stage-D matched-pair objective}
\label{app:stageD_training}
For a matched strong--weak pair $(X^{+},X^{-})$, let $y$ denote the
semantic answer and let $a=\mathrm{Abstain}$. We define the support
score for target $u$ as
\[
S(X,u)=-\operatorname{CE}\!\left(f(X),u\right),
\]
so that larger $S(X,u)$ indicates greater model support for $u$. With margin $m=0.5$, the matched-pair objective imposes four hinge
constraints:
\begin{align}
h_{\mathrm{strong}}
&=
\max\!\left(
0,\,
m - S(X^{+},y) + S(X^{+},a)
\right),\\
h_{\mathrm{weak}}
&=
\max\!\left(
0,\,
m - S(X^{-},a) + S(X^{-},y)
\right),\\
h_{\mathrm{sem}}
&=
\max\!\left(
0,\,
m - S(X^{+},y) + S(X^{-},y)
\right),\\
h_{\mathrm{abs}}
&=
\max\!\left(
0,\,
m - S(X^{-},a) + S(X^{+},a)
\right).
\end{align}
The first two terms enforce the desired prediction within each evidence
condition: the strong configuration should prefer the semantic answer
over abstention, whereas the weak configuration should prefer abstention
over the semantic answer. The latter two terms enforce the desired
change across the matched intervention: semantic-answer support should
be higher under strong evidence, while abstention support should be
higher under weak evidence.

The matched-pair loss is
\[
\mathcal{L}_{\mathrm{match}}
=
\frac{1}{4}
\left(
h_{\mathrm{strong}}
+
h_{\mathrm{weak}}
+
h_{\mathrm{sem}}
+
h_{\mathrm{abs}}
\right).
\]

The complete Stage-D objective is
\begin{equation}
\mathcal{L}_{\mathrm{pair\text{-}aware}}
=
\mathcal{L}_{\mathrm{ans+suff}}
+
\lambda_g \mathcal{L}_{\mathrm{ground}}
+
\lambda_m \mathcal{L}_{\mathrm{match}},
\end{equation}
with $\lambda_g=0.5$ and $\lambda_m=0.25$.
Stage D therefore tests whether direct coupling across matched
strong--weak predictions provides additional benefit beyond
sufficiency-aware answer supervision and referential grounding
supervision.

\subsection{Overall Gemma and Qwen Performance on 6-family questions}
\label{app:6fam_perfamily}
Tables~\ref{tab:gemma_6family_per_family_ans},~\ref{tab:gemma_6family_grounding}, ~\ref{tab:qwen_6family_per_family_ans}, and~\ref{tab:qwen_6family_grounding} report the complete six-family per-family results for Gemma and Qwen, respectively.

\paragraph{Gemma-4-12B per-family behavior.}
Gemma shows the clearest Stage B$\rightarrow$C grounding gains in Motion State,
Spatial Relation, and Temporal Relation. Motion State Joint@.50 rises from
60.46\% to 88.59\%, with JAGS@.50 increasing from 44.91\% to 73.22\%.
Spatial Relation improves from 79.98\% to 91.81\% Joint@.50 and from 66.82\%
to 82.72\% JAGS@.50, while Temporal Relation rises from 46.35\% to 78.54\%
Joint@.50 and from 8.90\% to 28.08\% JAGS@.50. Cross-view Recovery remains
the hardest family despite improvement in target-level grounding: Joint@.50
increases from 11.94\% at Stage B to 51.49\% at Stage C and 56.47\% at Stage D,
but All-G@.50 remains only 14.43\% and 21.39\% at the latter two stages.
Temporal Recovery shows a similar pattern at smaller scale, reaching only
34.48\% All-G@.50 at Stage D. Across most families, Stage D changes the
Stage-C result only modestly, consistent with the aggregate C--D comparison. Results are shown in Tables~\ref{tab:gemma_6family_per_family_ans} and~\ref{tab:gemma_6family_grounding}
\begin{table}[!ht]
\centering
\caption{Per-family Gemma-4-12B answering results on the canonical six-family held-out test set. Pair denotes matched-pair QA success. \textbf{All results are expressed as percentages}. Cross. is Cross-view Recovery; Map Rel. is Map Relation; Temp. Rec. is Temporal Recovery; Temp. Rel. is Temporal Relation.}
\label{tab:gemma_6family_per_family_ans}

\begin{tabular}{@{}lcrrrrr@{}}
\toprule
\textbf{Family} &
\textbf{Stage} &
\textbf{Strong $\uparrow$} &
\textbf{Weak $\uparrow$} &
\textbf{Pair $\uparrow$} &
\textbf{UCR $\downarrow$} &
\textbf{UAR $\downarrow$} \\
\midrule

\multirow{3}{*}{Cross-view Recovery}
& B & 100.00 & 100.00 & 100.00 & 0.00 & 0.00 \\
& C & 100.00 & 100.00 & 100.00 & 0.00 & 0.00 \\
& D & 100.00 & 100.00 & 100.00 & 0.00 & 0.00 \\
\midrule

\multirow{3}{*}{Map Relation}
& B & 50.81 & 98.38 & 50.27 & 1.62 & 10.27 \\
& C & 50.27 & 98.92 & 49.73 & 1.08 & 10.27 \\
& D & 49.19 & 98.38 & 48.65 & 1.62 & 8.65 \\
\midrule

\multirow{3}{*}{Motion State}
& B & 84.76 & 100.00 & 84.76 & 0.00 & 0.09 \\
& C & 86.65 & 99.73 & 86.38 & 0.27 & 0.18 \\
& D & 86.65 & 100.00 & 86.65 & 0.00 & 0.18 \\
\midrule

\multirow{3}{*}{Spatial Relation}
& B & 93.55 & 98.96 & 92.63 & 1.04 & 5.88 \\
& C & 93.32 & 98.50 & 92.05 & 1.50 & 5.65 \\
& D & 93.43 & 98.39 & 92.05 & 1.61 & 5.07 \\
\midrule

\multirow{3}{*}{Temporal Recovery}
& B & 100.00 & 100.00 & 100.00 & 0.00 & 0.00 \\
& C & 100.00 & 100.00 & 100.00 & 0.00 & 0.00 \\
& D & 100.00 & 100.00 & 100.00 & 0.00 & 0.00 \\
\midrule

\multirow{3}{*}{Temporal Relation}
& B & 52.74 & 100.00 & 52.74 & 0.00 & 1.60 \\
& C & 55.48 & 100.00 & 55.48 & 0.00 & 1.14 \\
& D & 57.76 & 100.00 & 57.76 & 0.00 & 0.91 \\
\midrule

\multirow{3}{*}{\textbf{Overall}}
& B & 81.52 & \textbf{99.58} & 81.20 & \textbf{0.42} & 2.76 \\
& C & 82.58 & 99.36 & 82.05 & 0.64 & 2.65 \\
& D & \textbf{82.90} & 99.40 & \textbf{82.44} & 0.60 & \textbf{2.44} \\

\bottomrule
\end{tabular}
\end{table}

\begin{table}[!ht]
\centering
\caption{
Per-family Gemma-4-12B grounding and evidence-supported answering results
on the canonical six-family held-out test set. Joint@.50 is target-level
joint presence--localization success; All-G@.50 requires all grounding
targets for a question to succeed; JAGS@.50 additionally requires Pair
Success. We report $t=0.50$.
}
\label{tab:gemma_6family_grounding}

\begin{tabular}{@{}lcrrr@{}}
\toprule
\textbf{Family} &
\textbf{Stage} &
\textbf{Joint@.50 (\%) $\uparrow$} &
\textbf{All-G@.50 (\%) $\uparrow$} &
\textbf{JAGS@.50 (\%) $\uparrow$} \\
\midrule

\multirow{3}{*}{Cross-view Recovery}
& B & 11.94 & 15.42 & 15.42 \\
& C & 51.49 & 14.43 & 14.43 \\
& D & 56.47 & 21.39 & 21.39 \\
\midrule

\multirow{3}{*}{Map Relation}
& B & 77.03 & 58.38 & 30.27 \\
& C & 90.54 & 82.16 & 42.16 \\
& D & 90.54 & 82.16 & 40.54 \\
\midrule

\multirow{3}{*}{Motion State}
& B & 60.46 & 53.56 & 44.91 \\
& C & 88.59 & 83.77 & 73.22 \\
& D & 89.04 & 84.67 & 73.94 \\
\midrule

\multirow{3}{*}{Spatial Relation}
& B & 79.98 & 69.01 & 66.82 \\
& C & 91.81 & 84.22 & 82.72 \\
& D & 91.81 & 84.45 & 82.26 \\
\midrule

\multirow{3}{*}{Temporal Recovery}
& B & 27.59 & 24.14 & 24.14 \\
& C & 60.34 & 27.59 & 27.59 \\
& D & 63.79 & 34.48 & 34.48 \\
\midrule

\multirow{3}{*}{Temporal Relation}
& B & 46.35 & 14.61 & 8.90 \\
& C & 78.54 & 49.54 & 28.08 \\
& D & 78.42 & 49.32 & 29.68 \\
\midrule

\multirow{3}{*}{\textbf{Overall}}
& B & 59.75 & 49.58 & 42.79 \\
& C & 84.41 & 73.00 & 62.37 \\
& D & \textbf{84.87} & \textbf{73.96} & \textbf{63.29} \\

\bottomrule
\end{tabular}
\end{table}

\paragraph{Qwen3-VL-8B per-family behavior.}
Qwen exhibits the same broad separation between selective answering and
grounding quality. At Stage B, Pair Success is already high for several
families, including 100.00\% for Cross-view Recovery, 89.09\% for Motion State,
and 90.90\% for Spatial Relation, while complete grounding is substantially
lower. Cross-view Recovery is the strongest example: Stage-B JAGS@.50 is only
1.00\%, increasing to 26.37\% at both Stages C and D. Motion State improves
from 37.96\% JAGS@.50 at Stage B to 72.59\% at Stage C and 73.94\% at Stage D,
while Spatial Relation rises from 68.32\% to 81.91\% and 82.49\%.
Temporal Relation also improves substantially in grounding, with Joint@.50
increasing from 46.23\% at Stage B to 76.60\% at Stage C, although JAGS@.50
remains lower at 23.29\%. Map Relation shows a smaller gain, and Temporal
Recovery remains difficult because of the small held-out population.
As with Gemma, most family-level changes from Stage C to Stage D are small
relative to the Stage B$\rightarrow$C grounding gains.%
\begin{table}[!t]
\centering
\caption{Per-family Qwen3-VL-8B answering results on the canonical six-family held-out test set. Pair denotes matched-pair QA success. \textbf{Results are expressed as percentages.} Cross. is Cross-view Recovery; Map Rel. is Map Relation; Temp. Rec. is Temporal Recovery; Temp. Rel. is Temporal Relation.}
\label{tab:qwen_6family_per_family_ans}

\begin{tabular}{@{}p{1.5cm}cccccc@{}}
\toprule
\textbf{Family} &
\textbf{Stage} &
\textbf{Strong (\%) $\uparrow$} &
\textbf{Weak (\%) $\uparrow$} &
\textbf{Pair (\%) $\uparrow$} &
\textbf{UCR (\%) $\downarrow$} &
\textbf{UAR (\%) $\downarrow$} \\
\midrule

\multirow{3}{*}{\parbox{1.5cm}{Cross-view\\Recovery}}
& B & 100.00 & 100.00 & 100.00 & 0.00 & 0.00 \\
& C & 100.00 & 100.00 & 100.00 & 0.00 & 0.00 \\
& D & 100.00 & 99.00 & 99.00 & 1.00 & 0.00 \\
\midrule

\multirow{3}{*}{Map Relation}
& B & 58.38 & 96.76 & 56.76 & 3.24 & 5.95 \\
& C & 58.92 & 96.22 & 56.76 & 3.78 & 5.95 \\
& D & 57.84 & 95.14 & 55.68 & 4.86 & 4.86 \\
\midrule

\multirow{3}{*}{Motion State}
& B & 89.09 & 100.00 & 89.09 & 0.00 & 0.00 \\
& C & 86.74 & 100.00 & 86.74 & 0.00 & 0.00 \\
& D & 87.92 & 100.00 & 87.92 & 0.00 & 0.00 \\
\midrule

\multirow{3}{*}{\parbox{1.5cm}{Spatial\\Relation}}
& B & 94.82 & 95.16 & 90.90 & 4.84 & 3.69 \\
& C & 94.59 & 95.97 & 91.24 & 4.03 & 3.69 \\
& D & 94.82 & 95.39 & 91.13 & 4.61 & 3.69 \\
\midrule

\multirow{3}{*}{\parbox{1.5cm}{Temporal\\Recovery}}
& B & 100.00 & 100.00 & 100.00 & 0.00 & 0.00 \\
& C & 100.00 & 100.00 & 100.00 & 0.00 & 0.00 \\
& D & 100.00 & 100.00 & 100.00 & 0.00 & 0.00 \\
\midrule

\multirow{3}{*}{\parbox{1.5cm}{Temporal\\Relation}}
& B & 53.20 & 100.00 & 53.20 & 0.00 & 0.91 \\
& C & 50.91 & 100.00 & 50.91 & 0.00 & 0.00 \\
& D & 51.60 & 100.00 & 51.60 & 0.00 & 0.00 \\
\midrule

\multirow{3}{*}{\textbf{Overall}}
& B & \textbf{84.17} & 98.30 & \textbf{82.86} & 1.70 & 1.66 \\
& C & 82.86 & \textbf{98.52} & 81.70 & \textbf{1.48} & 1.52 \\
& D & 83.43 & 98.20 & 82.08 & 1.80 & \textbf{1.45} \\

\bottomrule
\end{tabular}
\end{table}
\begin{table}[t]
\centering
\caption{Per-family Qwen3-VL-8B grounding and evidence-supported answering results on the canonical six-family held-out test set. Joint@.50 is target-level joint presence--localization success; All-G@.50 requires all grounding targets for a question to succeed; JAGS@.50 additionally requires Pair Success. We report $t=0.50$. 
}
\label{tab:qwen_6family_grounding}

\begin{tabular}{@{}lcccc@{}}
\toprule
\textbf{Family} &
\textbf{Stage} &
\textbf{Joint@.50 (\%) $\uparrow$} &
\textbf{All-G@.50 (\%) $\uparrow$} &
\textbf{JAGS@.50 (\%) $\uparrow$} \\
\midrule

\multirow{3}{*}{Cross-view Recovery}
& B & 12.44 & 1.00 & 1.00 \\
& C & 57.71 & 26.37 & 26.37 \\
& D & 57.96 & 26.37 & 26.37 \\
\midrule

\multirow{3}{*}{Map Relation}
& B & 77.30 & 55.68 & 34.59 \\
& C & 78.11 & 63.78 & 40.00 \\
& D & 78.11 & 63.78 & 38.92 \\
\midrule

\multirow{3}{*}{Motion State}
& B & 57.48 & 41.93 & 37.96 \\
& C & 87.15 & 83.59 & 72.59 \\
& D & 87.02 & 83.59 & 73.94 \\
\midrule

\multirow{3}{*}{Spatial Relation}
& B & 83.30 & 69.82 & 68.32 \\
& C & 91.37 & 84.10 & 81.91 \\
& D & 91.59 & 84.68 & 82.49 \\
\midrule

\multirow{3}{*}{Temporal Recovery}
& B & 27.59 & 13.79 & 13.79 \\
& C & 74.14 & 58.62 & 58.62 \\
& D & 75.86 & 58.62 & 58.62 \\
\midrule

\multirow{3}{*}{Temporal Relation}
& B & 46.23 & 9.59 & 4.57 \\
& C & 76.60 & 46.80 & 23.29 \\
& D & 76.48 & 47.49 & 22.83 \\
\midrule

\multirow{3}{*}{\textbf{Overall}}
& B & 59.67 & 43.18 & 39.01 \\
& C & 83.10 & 72.44 & 62.26 \\
& D & \textbf{83.11} & \textbf{72.72} & \textbf{62.83} \\

\bottomrule
\end{tabular}
\end{table}

Overall, the per-family results show that while some families gain stronger JAGS with training, some failure persisted, suggesting challenging families can improve in answering and grounding separately yet still struggle to satisfy both requirements simultaneously.

\subsection{Overall Gemma and Qwen Performance on 7-family set}
\label{app:7fam_adaptation}
The following results are reported from separate adaptation training with the Decision family included and should not be interpreted as equivalent to the results reported in the six-family study.

\textbf{Sufficiency and Grounding Supervision Drive Most Adaptation Gains.}
The adaptation results in Tables~\ref{tab:gemma-seven-family-ans} and~\ref{tab:qwen-seven-family-grounding} show that the largest change from zero-shot behavior comes from supervising evidence-conditioned answering. Both adapted backbones achieve high abstention accuracy and Pair Success while preserving answering accuracy under supported condition. Here, Gemma-C reaches 81.58\% Strong Accuracy, 98.93\% Weak Abstention Accuracy, and 80.68\% Pair Success. Qwen-C shows the same overall pattern at a slightly higher QA level, reaching 83.95\% Strong Accuracy, 98.53\% Weak Abstention Accuracy, and 82.78\% Pair Success. Thus, evidence-sensitive answering is not specific to a single backbone.

Grounding is evaluated through a prediction of presence and, when found, observation index and normalized box. Construction boxes are hidden from the model and used only for scoring. On the test set, the Gemma-C model reaches 96.82\% presence accuracy, 76.67\% localization accuracy at IoU $\geq 0.5$, and 77.16\% joint presence-\-localization success. Stage D changes these aggregate grounding results minimally, reaching 97.03\% presence accuracy, 76.76\% IoU $\geq0.5$, and 77.43\% Joint@0.5. Similarly, Qwen-C reaches 96.45\% presence accuracy, 74.01\%
localization accuracy at IoU $\geq 0.5$, and 74.71\% joint
presence--localization success. Stage D is similar, with 96.49\%
presence accuracy, 73.70\% localization accuracy at IoU $\geq 0.5$,
and 74.46\% joint presence--localization success. Thus, these results show that grounding supervision substantially improves the adapted models' localization capabilities and that the additional Stage D pair objective does not produce a uniform aggregate localization improvement beyond explicit grounding supervision. Despite their differences in their zero-shot behavior, the two backbones converge to similar evidence-supported performance after adaptation. Qwen retains slightly stronger strong-condition understanding, whereas Gemma attains slightly stronger grounding and question-level evidence completion.

\textbf{Explicit matched-pair objective Does Not Produce Uniform Improvements.} The explicit matched-pair ranking objective in Stage D has a family-dependent effect. For Gemma, Stage D slightly increases Joint@0.5 from 77.16\% to 77.43\% and All-G@0.5 from 72.58\% to 73.19\%, but Strong Accuracy falls from 81.58\% to 78.87\%, Pair Success from 80.68\% to 78.13\%, and JAGS@0.5 from 61.79\% to 59.41\%. Qwen shows an even smaller C-D difference: Pair Success changes from 82.78\% to 82.25\%, Joint@0.5 from 74.71\% to 74.45\%, and JAGS@0.5 from  61.12\% to 60.62\%. These results suggest that explicit pair coupling is not a universally additive improvement once sufficiency and grounding supervision are already present; its effect is small and can vary by backbone and question family.  Appendix~\ref{app:bootstrap} presents paired bootstrap uncertainty. Detailed per-family results for Gemma and Qwen are reported below in Appendix~\ref{app:qwen_detailed}.

\subsubsection{Per-family Gemma Performance}

\textbf{Adapted Model Performance Varies by Reasoning Family} Tables~\ref{tab:gemma-seven-family-ans} and~\ref{tab:gemma-seven-family-grounding} show the differences in performance of all models across the different families. For both backbones, spatial and motion questions are among the strongest evidence-supported categories. Gemma-C reaches 81.91\% JAGS@0.5 on spatial relation and 75.02\% on motion state, while Qwen reaches 83.64\% and 75.38\%, respectively as seen in Table~\ref{tab:qwen-seven-family-grounding}. In contrast, Decision remains substantially harder: JAGS@0.5 is only 12.42\% for Gemma C and 13.66\% for Qwen-C, despite much higher Pair Success. This gap suggests that producing appropriate answer-abstain behavior is easier than grounding all of the policy-relevant actors required to support a driving decision. Map Relation reveals another challenge: JAGS@0.5 near 38--39\% for both backbones, while
temporal relation remains more difficult at 31.28\% for Gemma-C and 24.43\%
for Qwen-C. The family-level comparison thus reveals the differences and challenges that each family poses for the adapted models.

\begin{table}[!ht]
\centering

\caption{
Per-family Gemma-4-12B answering results on the full seven-family
held-out test set. Pair denotes matched-pair QA success.}
\label{tab:gemma-seven-family-ans}

\begin{tabular}{@{}lcccc@{}}
\toprule
\textbf{Family} &
\textbf{Stage} &
\textbf{Strong (\%) $\uparrow$} &
\textbf{Weak (\%) $\uparrow$} &
\textbf{Pair (\%) $\uparrow$} \\
\midrule

\multirow{2}{*}{Cross-view}
& C & 100.00 & 99.50 & 99.50 \\
& D & 100.00 & 100.00 & 100.00 \\
\midrule

\multirow{2}{*}{Decision}
& C & 63.35 & 100.00 & 63.35 \\
& D & 67.70 & 98.14 & 66.46 \\
\midrule

\multirow{2}{*}{Map Relation}
& C & 54.05 & 98.38 & 53.51 \\
& D & 52.43 & 98.38 & 51.89 \\
\midrule

\multirow{2}{*}{Motion State}
& C & 87.20 & 99.37 & 86.56 \\
& D & 78.09 & 99.82 & 77.91 \\
\midrule

\multirow{2}{*}{Spatial Relation}
& C & 94.12 & 97.58 & 92.05 \\
& D & 94.70 & 97.58 & 92.74 \\
\midrule

\multirow{2}{*}{Temporal Recover}
& C & 100.00 & 100.00 & 100.00 \\
& D & 100.00 & 100.00 & 100.00 \\
\midrule

\multirow{2}{*}{Temporal Relation}
& C & 51.14 & 100.00 & 51.14 \\
& D & 53.65 & 100.00 & 53.65 \\
\midrule

\multirow{2}{*}{\textbf{Overall}}
& C & \textbf{81.58} & 98.93 & \textbf{80.68} \\
& D & 78.87 & \textbf{99.03} & 78.13 \\

\bottomrule
\end{tabular}
\end{table}

\begin{table}[!ht]
\centering

\caption{
Per-family Gemma-4-12B grounding and evidence-supported answering results
on the full seven-family held-out test set. Joint@.50 is target-level
joint presence--localization success; All-G@.50 requires all grounding
targets associated with a question to succeed; JAGS@.50 additionally
requires matched-pair QA success. We report $t=0.50$. 
}
\label{tab:gemma-seven-family-grounding}

\begin{tabular}{@{}lcrrr@{}}
\toprule
\textbf{Family} &
\textbf{Stage} &
\textbf{Joint@.50 (\%) $\uparrow$} &
\textbf{All-G@.50 (\%) $\uparrow$} &
\textbf{JAGS@.50 (\%) $\uparrow$} \\
\midrule

\multirow{2}{*}{Cross-view Recovery}
& C & 59.20 & 29.35 & 29.35 \\
& D & 60.20 & 29.85 & 29.85 \\
\midrule

\multirow{2}{*}{Decision}
& C & 26.94 & 26.09 & 12.42 \\
& D & 26.94 & 27.33 & 13.04 \\
\midrule

\multirow{2}{*}{Map Relation}
& C & 85.95 & 74.59 & 39.46 \\
& D & 85.68 & 72.97 & 37.84 \\
\midrule

\multirow{2}{*}{Motion State}
& C & 89.59 & 85.93 & 75.02 \\
& D & 90.44 & 87.29 & 67.54 \\
\midrule

\multirow{2}{*}{Spatial Relation}
& C & 90.82 & 83.29 & 81.91 \\
& D & 91.81 & 85.37 & 84.10 \\
\midrule

\multirow{2}{*}{Temporal Recovery}
& C & 75.86 & 55.17 & 55.17 \\
& D & 62.07 & 34.48 & 34.48 \\
\midrule

\multirow{2}{*}{Temporal Relation}
& C & 79.97 & 54.79 & 31.28 \\
& D & 79.34 & 52.74 & 31.28 \\
\midrule

\multirow{2}{*}{\textbf{Overall}}
& C & 77.16 & 72.58 & \textbf{61.79} \\
& D & \textbf{77.43} & \textbf{73.19} & 59.41 \\

\bottomrule
\end{tabular}
\end{table}

\subsubsection{Per-family Qwen performance}
\label{app:qwen_detailed}
Qwen exhibits the same broad pattern as Gemma: the effect of Stage D matched-pair objective is family-dependent rather than uniform improvements. Cross-view Recovery and Temporal Recovery remain high in matched-pair QA, while Spatial Relation and Temporal Relation remain substantially harder, with lower Pair Success. Table~\ref{tab:qwen-7family-ans} illustrates these detailed results.

Grounding shows a different pattern as seen in Table~\ref{tab:qwen-seven-family-grounding}. Qwen achieves high target-level localization on Spatial Relation, Motion State, Map Relation, but complete-evidence recovery is much weaker for Cross-view Recovery and Decision. This gap between target-level grounding and question-level All-G/JAGS indicates that Qwen can often localize individual targets without consistently recovering the full evidence set required for a matched question. For Stage~C, JAGS@.50 ranges from 13.66\% on Decision and 25.87\% on
Cross-View Recovery to 75.38\% on Motion State and 83.64\% on Spatial Relation,
highlighting substantial variation in how reliably Qwen combines correct
answering with complete evidence recovery.

Stage D produces only modest changes overall with Spatial Relation remaining strong while Cross-view Recovery and Temporal Recovery experiencing mixed results. This result reinforces the finding that explicit matched-pair coupling does not yield a monotonic improvement across different models (Gemma and Qwen).
\begin{table}[!ht]
\centering
\caption{Per-family QA results for adapted Qwen3-VL-8B on the
seven-family held-out test set. Cross. is Cross-view Recovery; Decision is Evidence-conditioned Decision; Map is Map Relation; Motion is Motion State; Spatial is Spatial Relation; Temp. Rec. is Temporal Recovery; Temp. Rel. is Temporal Relation.}
\label{tab:qwen-7family-ans}
{%
\begin{tabular}{p{1.5cm}cccccc}
\toprule
\textbf{Family} & \textbf{Stage} &
\textbf{Strong} (\%) $\uparrow$ &
\textbf{Weak} (\%) $\uparrow$ &
\textbf{Pair} (\%) $\uparrow$ &
\textbf{UCR} (\%) $\downarrow$ &
\textbf{UAR} (\%) $\downarrow$ \\
\midrule
\multirow{2}{*}{\parbox{1.5cm}{Cross.}}
& C & 100.00 & 100.00 & 100.00 & 0.00 & 0.00 \\
& D & 100.00 & 100.00 & 100.00 & 0.00 & 0.00 \\
\midrule
\multirow{2}{*}{Decision}
& C & 75.78 & 100.00 & 75.78 & 0.00 & 0.00 \\
& D & 76.40 & 100.00 & 76.40 & 0.00 & 0.00 \\
\midrule
\multirow{2}{*}{\parbox{1.5cm}{Map}}
& C & 57.84 & 95.68 & 55.68 & 4.32 & 5.41 \\
& D & 56.22 & 95.14 & 53.51 & 4.86 & 4.32 \\
\midrule
\multirow{2}{*}{Motion}
& C & 89.45 & 100.00 & 89.45 & 0.00 & 0.00 \\
& D & 88.37 & 100.00 & 88.37 & 0.00 & 0.00 \\
\midrule
\multirow{2}{*}{\parbox{1.5cm}{Spatial}}
& C & 95.05 & 95.85 & 91.47 & 4.15 & 3.57 \\
& D & 95.39 & 95.51 & 91.47 & 4.49 & 3.23 \\
\midrule
\multirow{2}{*}{\parbox{1.5cm}{Temp.\\Rec.}}
& C & 100.00 & 100.00 & 100.00 & 0.00 & 0.00 \\
& D & 100.00 & 100.00 & 100.00 & 0.00 & 0.00 \\
\midrule
\multirow{2}{*}{\parbox{1.5cm}{Temp.\\Rel.}}
& C & 53.65 & 100.00 & 53.65 & 0.00 & 0.68 \\
& D & 53.42 & 100.00 & 53.42 & 0.00 & 0.00 \\
\midrule
\multirow{2}{*}{\textbf{Overall}}
& C & \textbf{83.95} & \textbf{98.53} & \textbf{82.78} & \textbf{1.47} & 1.47 \\
& D & 83.55 & 98.40 & 82.25 & 1.60 & \textbf{1.20} \\
\bottomrule
\end{tabular}%
}
\end{table}

\begin{table}[!ht]
\centering
\caption{
Per-family grounding and evidence-supported answering for
Qwen3-VL-8B on the seven-family held-out test set.
Joint@.50 is target-level joint presence--localization success;
All-G@.50 requires all grounding targets associated with a question to
succeed; JAGS@.50 additionally requires matched-pair QA success. We report $t=0.50$. Cross. is Cross-view Recovery; Decision is Evidence-conditioned Decision; Map is Map Relation; Motion is Motion State; Spatial is Spatial Relation; Temp. Rec. is Temporal Recovery; Temp. Rel. is Temporal Relation. \textbf{All results are expressed as percentage.}
}
\label{tab:qwen-seven-family-grounding}
{
\begin{tabular}{lcrrrrr}
\toprule
\textbf{Family} & \textbf{Stage} &
\textbf{Presence} $\uparrow$ &
\textbf{IoU@$t$} $\uparrow$ &
\textbf{Joint@$t$} $\uparrow$ &
\textbf{All-G@$t$} $\uparrow$ &
\textbf{JAGS@$t$} $\uparrow$ \\
\midrule

Cross.
& C & 87.56 & 67.05 & 58.71 & 25.87 & 25.87 \\
& D & 88.06 & 66.38 & 58.46 & 26.87 & 26.87 \\

\midrule
Decision
& C & 98.77 & 21.99 & 22.31 & 25.47 & 13.66 \\
& D & 98.77 & 22.78 & 23.16 & 24.22 & 13.04 \\

\midrule
Map Rel.
& C & 84.86 & 85.00 & 77.57 & 62.70 & 37.84 \\
& D & 85.68 & 82.78 & 77.30 & 62.70 & 38.92 \\

\midrule
Motion state
& C & 98.92 & 88.15 & 87.20 & 83.41 & 75.38 \\
& D & 99.10 & 87.67 & 86.88 & 83.32 & 74.21 \\

\midrule
Spatial Rel.
& C & 95.58 & 92.92 & 92.09 & 85.83 & 83.64 \\
& D & 94.80 & 92.42 & 91.10 & 84.22 & 82.26 \\

\midrule
Temp. Rec.
& C & 94.83 & 76.36 & 72.41 & 51.72 & 51.72 \\
& D & 94.83 & 76.36 & 72.41 & 55.17 & 55.17 \\

\midrule
Temp. Rel.
& C & 97.37 & 77.78 & 75.74 & 45.89 & 24.43 \\
& D & 97.83 & 77.36 & 75.68 & 46.35 & 25.80 \\

\midrule
\textbf{Overall}
& C & 96.45 & \textbf{74.01} & \textbf{74.71}
& \textbf{70.04} & \textbf{61.12} \\
& D & \textbf{96.49} & 73.70 & 74.46
& 69.64 & 60.62 \\

\bottomrule
\end{tabular}%
}
\end{table}

\subsection{All-G as a function of target count}
All-G is an all-or-nothing question-level conjunction and therefore
becomes increasingly difficult as the number of required grounding
targets grows. The test set contains 2,429 questions with two
grounding targets and 506 questions with four. Only 56 questions
contain six or more grounding targets. Across all systems, All-G@.50 decreases substantially as target count
increases. For example, Gemma Stage~C falls from 79.09\% with two
targets to 49.41\% with four, while Qwen Stage~C falls from 77.73\%
to 40.91\%. The same pattern is substantially more severe in the
zero-shot models (see Table~\ref{tab:allg-target-count}). No evaluated system achieves All-G@.50 on the 56 questions requiring
six or more targets. All-G is lower for questions with more targets, although target count also varies with reasoning families and difficulty.
\begin{table}[!ht]
\centering
\caption{
All-G@$t$ by number of grounding targets. $t=0.50$. The two largest buckets contain 2,429 and 506 questions, respectively.
}
\label{tab:allg-target-count}
\begin{tabular}{lrr}
\toprule
\textbf{System} & \textbf{2 targets (\%) $\uparrow$}& \textbf{4 targets (\%) $\uparrow$}\\
\midrule
Gemma-C & 79.09 & 49.41 \\
Gemma-D & 80.16 & 47.83 \\
Qwen-C & 77.73 & 40.91 \\
Qwen-D & 77.15 & 41.30 \\
Gemma zero-shot & 48.70 & 8.50 \\
Qwen zero-shot & 54.43 & 8.50 \\
InternVL zero-shot & 0.21 & 0.00 \\
Molmo zero-shot & 0.04 & 0.00 \\
LLaVA zero-shot & 0.33 & 0.00 \\
\bottomrule
\end{tabular}
\end{table}

\section{Qualitative Examples and Failure Modes}
\label{app:qualitative}
This section provides qualitative examples for six scene-understanding families. \textbf{All examples use the original nuScenes~\citep{caesar2020nuscenes} camera frames; DriveAnswer
introduces no blur, degradation, or other image-quality modification.}

\subsection{Spatial and Map Reasoning}
\begin{figure*}[!t]
\centering
    \centering
    \includegraphics[width=\linewidth]{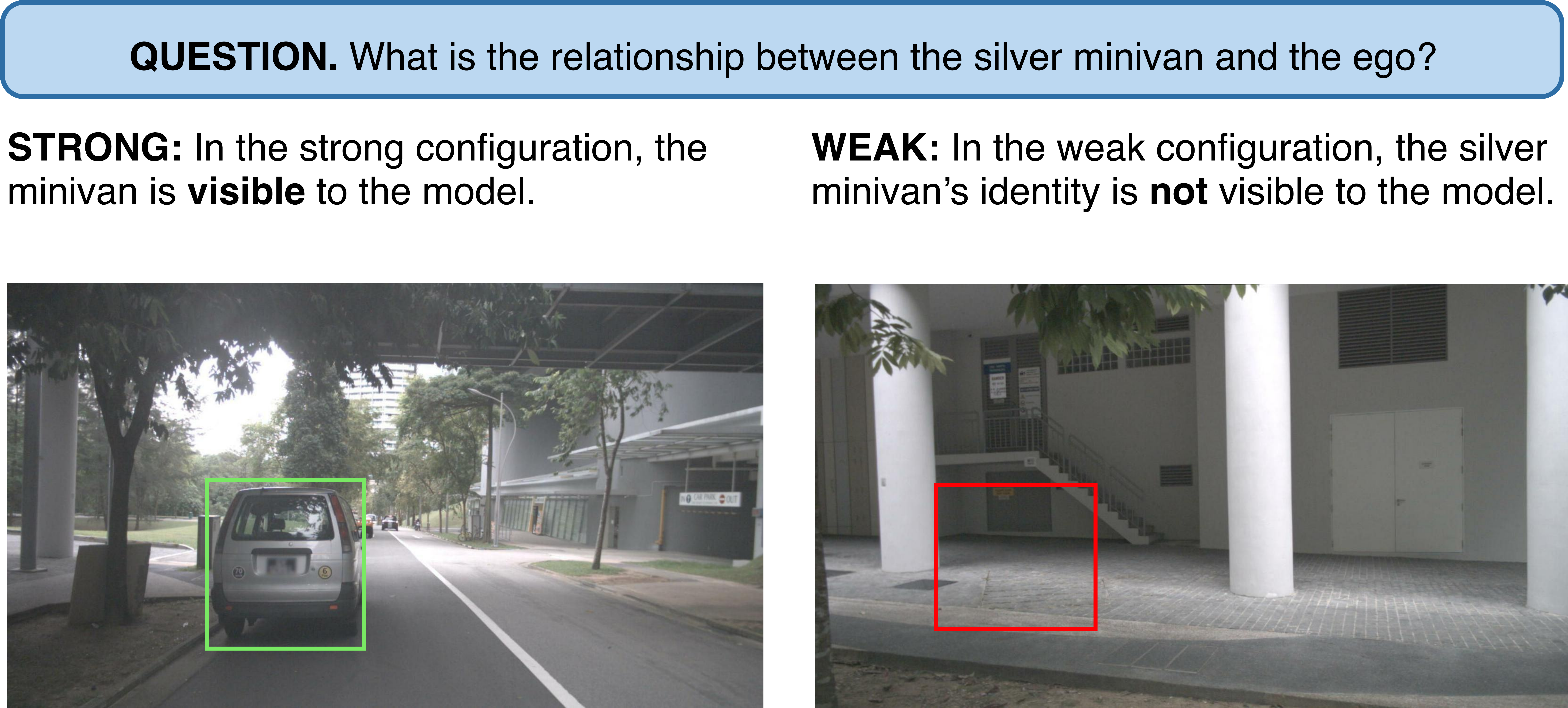}
        \caption{\textbf{Spatial Relation example.}
The strong configuration preserves the visual support needed to establish the target's relation to the ego, whereas the weak configuration withholds critical target evidence while preserving the underlying scene fact.}
    \label{fig:qual_spatial_reasoning}
\end{figure*}

\begin{figure*}[!t]
    \centering
    \includegraphics[width=\linewidth]{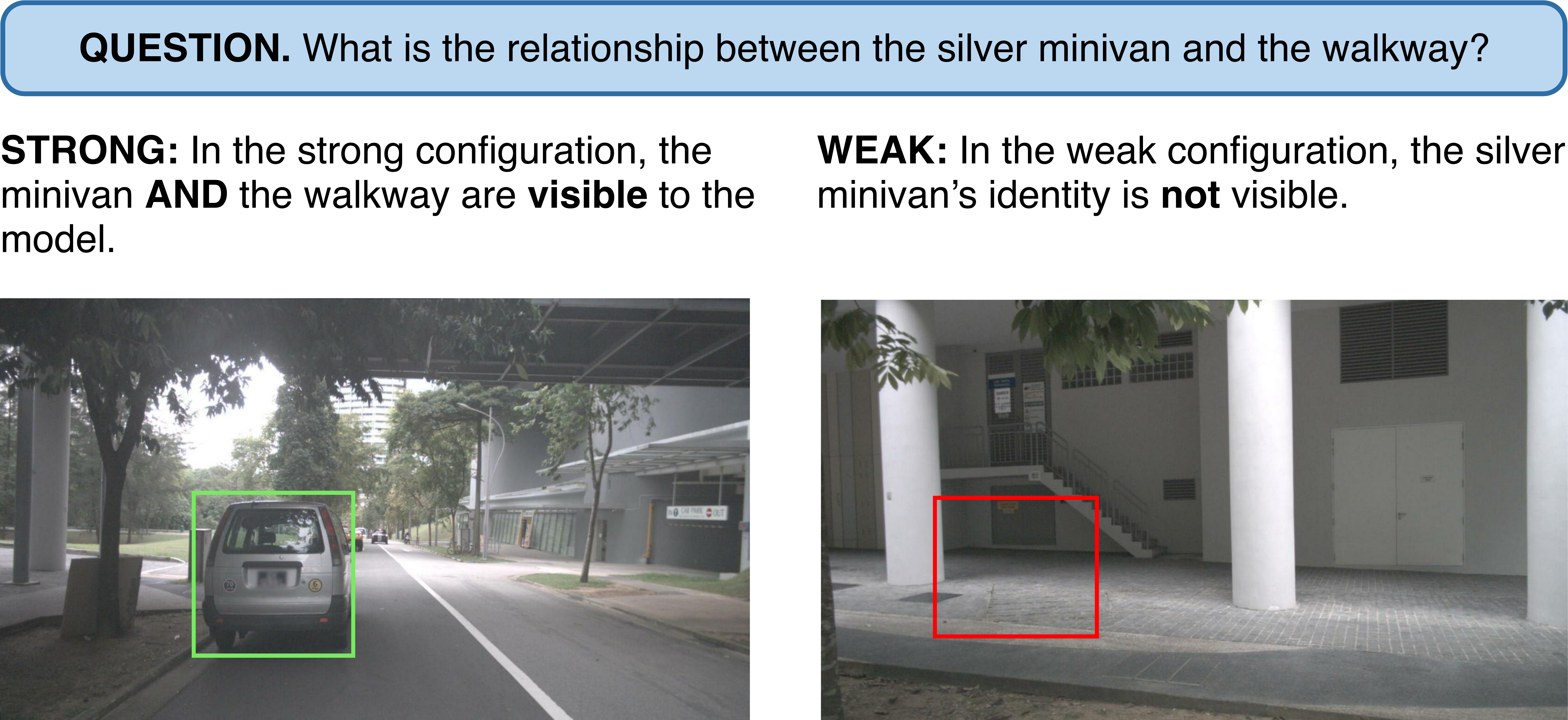}
        \caption{\textbf{Map Relation example.}
The strong configuration jointly supports the queried actor and its relation to mapped scene structure, whereas the weak configuration removes critical visual support required to establish that relation.}
    \label{fig:qual_map_reasoning}
\end{figure*}

Spatial and map examples illustrate that preserving an object in view
does not necessarily preserve the evidence required to establish its
relationship to another entity or to mapped scene structure. These
examples as illustrated in Figures~\ref{fig:qual_spatial_reasoning} and~\ref{fig:qual_map_reasoning} are useful for distinguishing target visibility
from relation sufficiency.

These examples illustrate two distinct recover mechanisms. Cross-view Recovery restores evidence through a synchronized alternative camera, whereas Temporal Recovery restores evidence through neighboring time steps. in both cases, the anchor observation alone is insufficient for question answering, so successful behavior requires using the additional admitted evidence defined by the each family.

The temporal examples further show why object visibility alone is insufficient. The queried actor may remain identifiable in the weak configuration, while the sequence no longer support answers about its motion or the evolution of its relation to another actor. 
\FloatBarrier

\subsection{Cross-view and Temporal Recovery}
\begin{figure*}[!t]
    \centering
    \includegraphics[width=\linewidth]{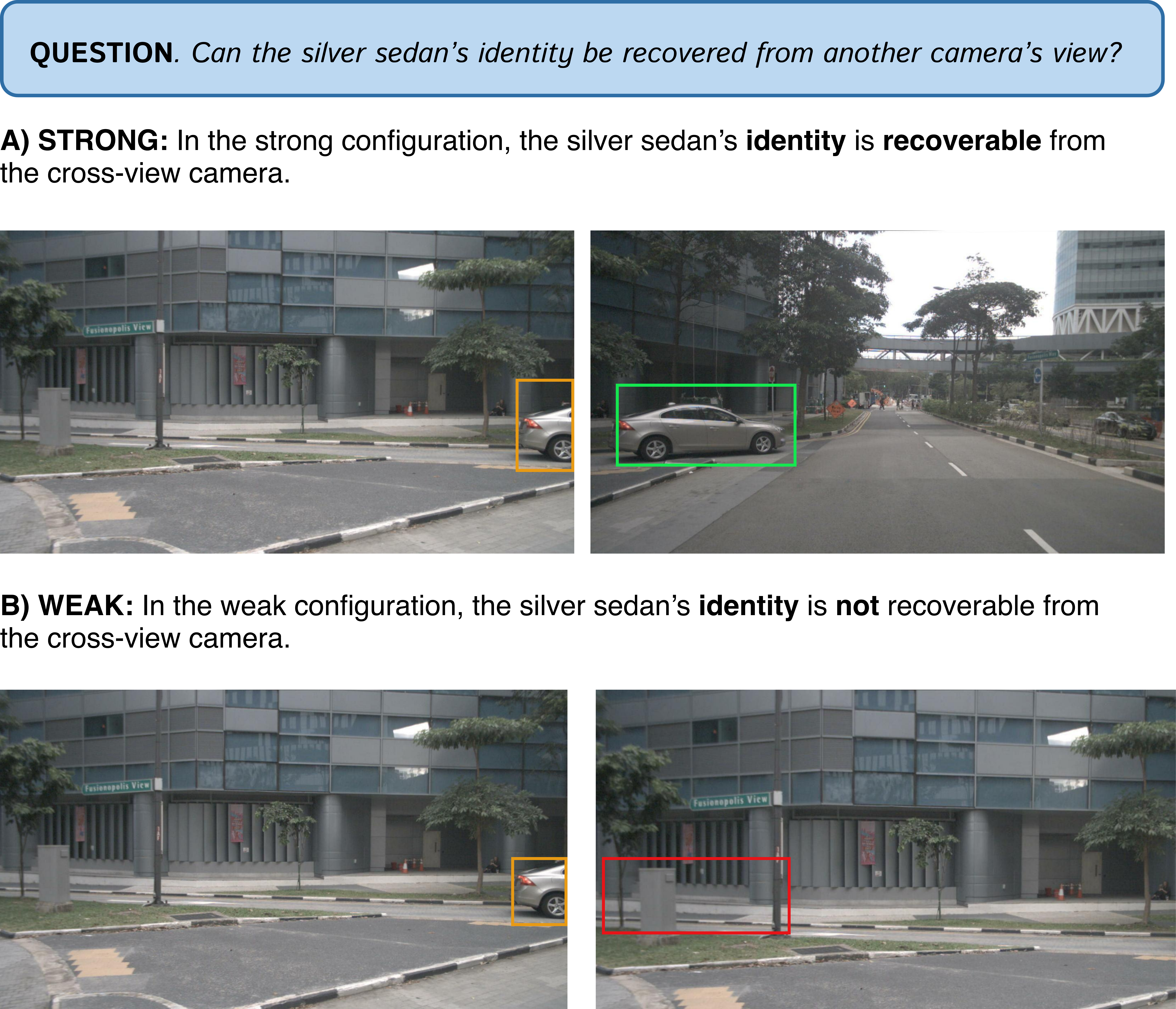}
    \caption{\textbf{Cross-view Recovery example.} The anchor observation alone is insufficient; the strong configuration provides synchronized camera view from which the queried actor can be recovered, whereas the weak sequence does not.
    \label{fig:qual_crossview_recovery}
}
\end{figure*}

\begin{figure}[!t]
    \centering
    \includegraphics[width=\linewidth]{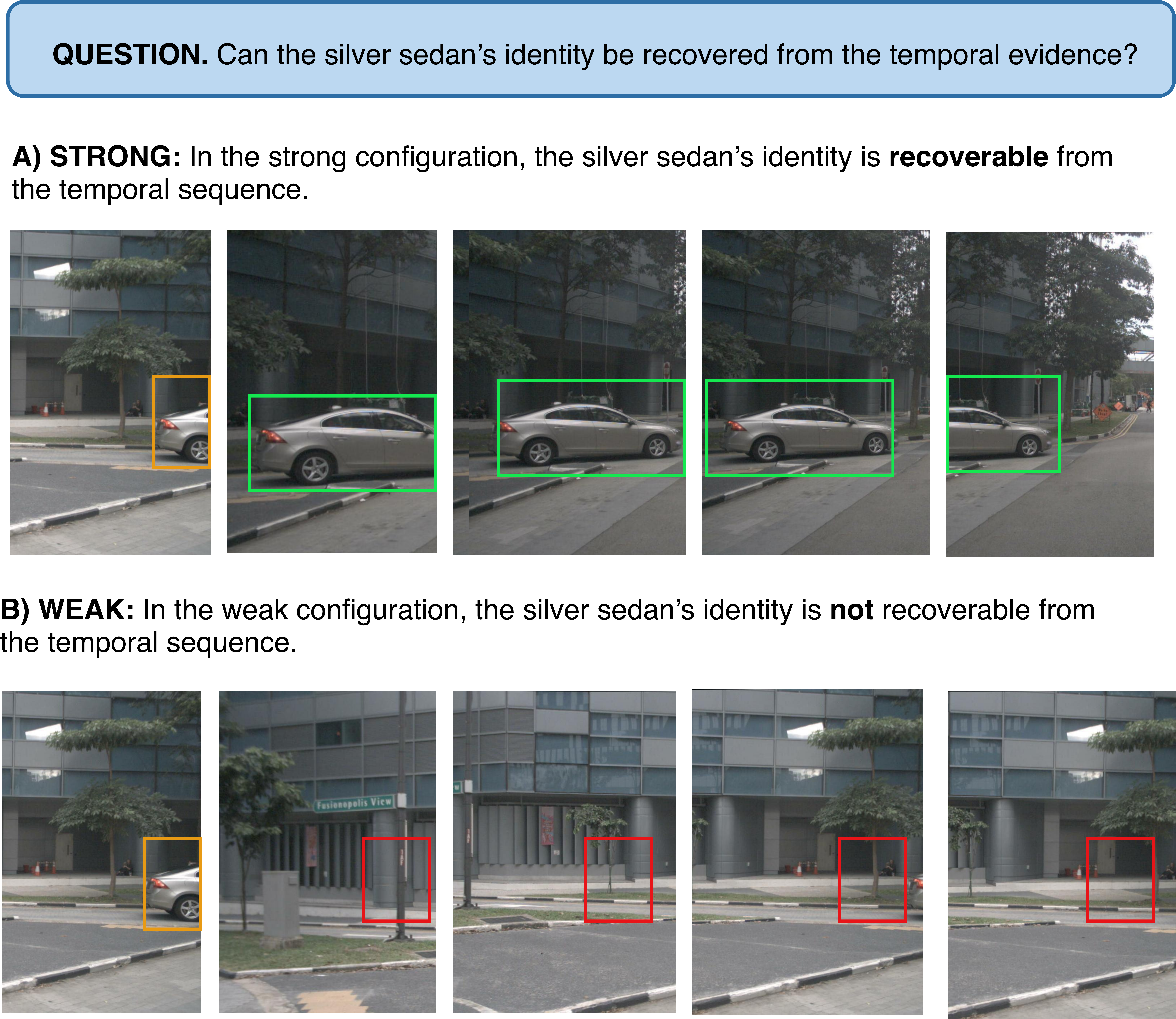}
    \caption{\textbf{Temporal Recovery example.} The anchor observation alone is insufficient; the strong configuration provides neighboring temporal observations that recover the missing evidence, whereas the weak sequence does not.}
    \label{fig:qual_temporal_recovery}
\end{figure}

Recovery examples require that the designated anchor observation be
insufficient by itself. Cross-view recovery adds a synchronized camera
that restores the missing evidence, while temporal recovery adds a
nearby time step. Examples in which the anchor already determines the
answer are treated as controls rather than as genuine recovery cases as seen in Figures~\ref{fig:qual_crossview_recovery} and~\ref{fig:qual_temporal_recovery}.

These recovery families test whether the model can use additional admitted observations to reconstruct evidence that is unavailable at the anchor frame. Specifically, Cross-view Recovery restores the missing evidence through synchronized alternative camera(s), whereas Temporal Recovery restores it through consecutive timesteps. The key distinction is that the semantic question and underlying scene are held fixed while the observation set changes. These families primarily focuses on whether the model can make use of additional observations when the evidence required by the question is unavailable at the anchor. A successful model should therefore answer only when the added cross-view or temporal observation restores the evidence needed for the queried inference, rather than relying on the anchor frame alone. These cases separate object visibility at a single observation from the broader question of whether the admitted observation set is sufficient to support the requested inference.

\FloatBarrier

\subsection{Motion State and Temporal Reasoning}
\begin{figure}[!t]
\centering
\includegraphics[width=\linewidth]{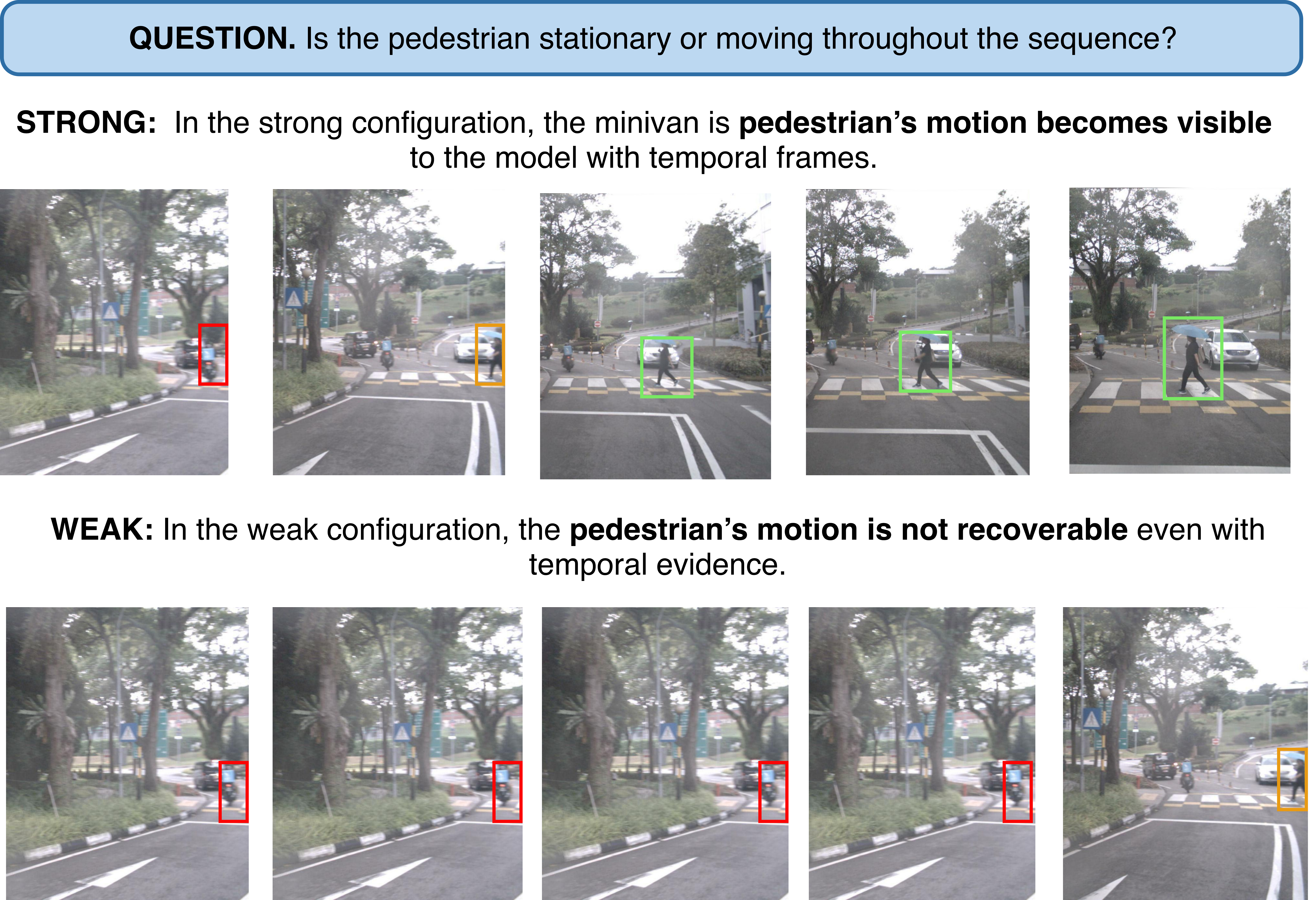}
\caption{\textbf{Motion State example.} The strong sequence provides sufficient temporal evidence to infer the queried actor's motion, whereas the weak sequence preserves partial observations without enough temporal support to determine its motion state.}
\label{fig:qual_motion}
\end{figure}

\begin{figure}[!tbp]
    \includegraphics[width=\linewidth]{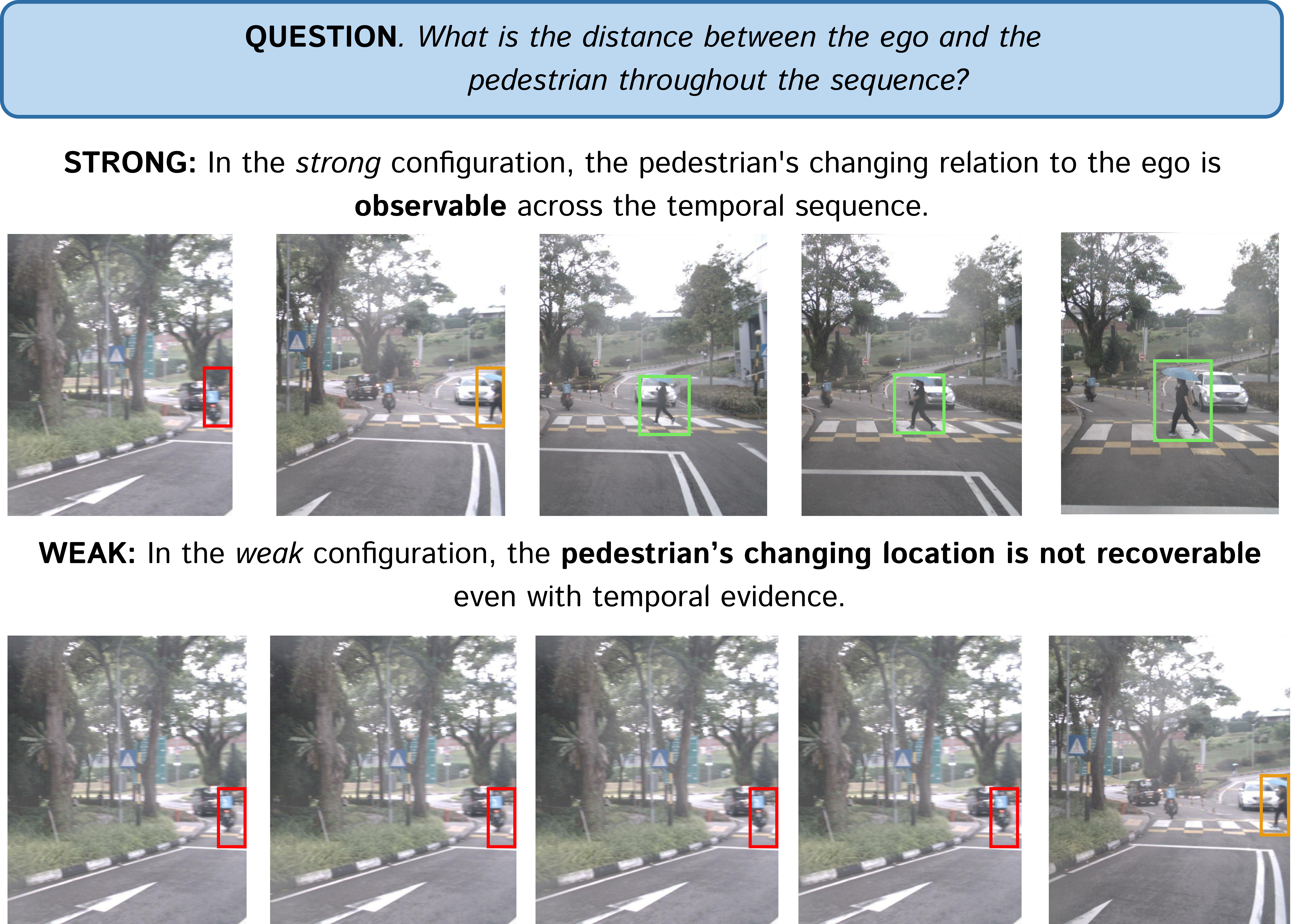}
\caption{\textbf{Temporal Relation example.} The strong sequence reveals how the queried relation changes over time, whereas the weak sequence lacks sufficient temporal evidence to establish that change.
}
\label{fig:qual_temporal_relation}
\end{figure}

Motion-state and temporal-relation examples use temporally separated
observations of the same scene. A single clear anchor image may be
sufficient to recognize an object while remaining insufficient to infer
whether it is moving or how its relation to another actor changes over
time. In qualitative visualizations in Figures~\ref{fig:qual_motion} and~\ref{fig:qual_temporal_relation}, the anchor frame may be displayed
at full emphasis and neighboring temporal frames at reduced visual
emphasis to make the temporal evidence structure explicit.

Motion state and temporal relation use temporally separated observations of the same scene. A single clear anchor image may be sufficient to recognize an actor while remaining insufficient to infer whether it is moving or how its relation to another actor changes over time. These temporal families are intentionally different from recovery in that the relevant actor may remain visible throughout the sequence, while the requested property cannot be established from appearance alone. Motion State requires evidence of how an actor changes across time, and Temporal Relation requires evidence of how the relation between actors evolve. The weak configurations therefore preserve partial visual content while removing enough temporal support to make the requested inference cannot be reliably answered. This distinction is important because a model can correctly identify the objects involved while still lacking the evidence needed to infer motion or temporal change. These examples therefore isolate a different failure mode from pure detection or presence errors. The benchmark asks whether the model conditions its answer on the temporal evidence actually available, rather than whether it can recognize the entities appearing in individual frames.

Taken together, the qualitative examples show that insufficient evidence does not necessarily correspond to an absent object or an empty observation. Across the six families, the strong-weak intervention primarily changes the evidence required for the queried inference rather than simply removing the required object. The examples therefore illustrate the benchmarks' central distinction between observing relevant content and having sufficient evidence to justify an answer. Conversely, the strong condition adds only the evidence needed to support the same underlying semantic question. These examples visualize the core benchmark distinction between observing relevant scene content and having sufficient evidence to support the requested inference.
\FloatBarrier